\pdfoutput=1
\documentclass[10pt,twocolumn,letterpaper]{article}

\usepackage[pagenumbers]{cvpr} %

\usepackage{graphicx}
\usepackage{amsmath,amsfonts,amssymb}
\usepackage{booktabs}
\usepackage[loop,autoplay]{animate}
\newif\ifcompileanimated %
\compileanimatedtrue  %

\ifcompileanimated
\else %
    \usepackage[accsupp]{axessibility}  %
\fi

\usepackage{siunitx}
\usepackage{adjustbox}
\usepackage{makecell}
\usepackage{multirow}
\usepackage{enumitem} %

\usepackage{pifont}%
\newcommand{\yesmark}{\textcolor{green}{\ding{51}}}%
\newcommand{\nomark}{\textcolor{red}{\ding{55}}}%

\usepackage{url}
\usepackage{titling}
\date{}\predate{}\postdate{}
\pretitle{\begin{center}\Large \bf}
\posttitle{\end{center}\vskip 0.5em}

\usepackage{color} 

\global\long\def\rpgkitchen{\emph{kitchen}}
\global\long\def\rpgsnoopy{\emph{snoopy}}

\global\long\def\rpgbin{\emph{bin}}
\global\long\def\rpgboxes{\emph{boxes}}
\global\long\def\rpgdesk{\emph{desk}}
\global\long\def\rpgmonitor{\emph{monitor}}

\global\long\def\edsfloorloop{\emph{04\_floor\_loop}}
\global\long\def\edsrpgbuilding{\emph{05\_rpg\_building}}
\global\long\def\edsflooreightloop{\emph{12\_floor\_eight\_loop}}
\global\long\def\edsaptday{\emph{15\_apartment\_day}}

\global\long\def\Lum{L}
\global\long\def\pol{p} %

\global\long\def\cE{\mathcal{E}}
\global\long\def\cT{\mathcal{T}}
\global\long\def\numEvents{N_e}

\global\long\def\bu{\mathbf{u}}

\global\long\def\pol{p}
\global\long\def\velflow{\mathbf{v}}
\global\long\def\linvel{\mathbf{V}}
\global\long\def\angvel{\boldsymbol{\omega}}

\global\long\def\Real{\mathbb{R}}
\global\long\def\cF{\mathcal{F}}

\RequirePackage{isomath}
\renewcommand{\vec}{\vectorsym}
\newcommand{\mat}{\matrixsym}

\usepackage{mathtools}

\newcommand{\deltapose}[0]{\delta\vec{T}}
\newcommand{\pose}[0]{\vec{T}}
\newcommand{\vel}[0]{\dot{\pose}}

\def\depthu{d_{\bu}}

\newcommand\scalemath[2]{\scalebox{#1}{\mbox{\ensuremath{\displaystyle #2}}}}

\newcommand\wframe[1]{\color{white}\frame{#1}}

\definecolor{somegray}{rgb}{0.5, 0.5, 0.5}
\newcommand{\darkgrayed}[1]{\textcolor{somegray}{#1}}
\makeatletter
\newcommand*\titleheader[1]{\gdef\@titleheader{#1}}
\AtBeginDocument{%
  \let\st@red@title\@title
  \def\@title{%
    \vskip-3.45em
    \bgroup\normalfont\large\centering\@titleheader\par\egroup
    \vskip1.5em\st@red@title}
}
\makeatother

\titleheader{\darkgrayed{This paper has been accepted for publication at the\\ 
IEEE Conference on Computer Vision and Pattern Recognition (CVPR), New Orleans, 2022.
\copyright IEEE}}

\def\MYTITLE{Event-aided Direct Sparse Odometry\vspace{-3ex}}

\usepackage[breaklinks,colorlinks]{hyperref}

\usepackage[capitalize]{cleveref}
\crefname{section}{Sec.}{Secs.}
\Crefname{section}{Section}{Sections}
\Crefname{table}{Table}{Tables}
\crefname{table}{Tab.}{Tabs.}

\def\cvprPaperID{2094}
\def\confName{CVPR}
\def\confYear{2022}

\title{\MYTITLE}

\author{Javier Hidalgo-Carri\'o$^{1}$, Guillermo Gallego$^{2}$, Davide Scaramuzza$^1$\\
$^{1}$Dept. of Informatics, Univ. of Zurich and Dept. of Neuroinformatics, Univ. of Zurich and ETH Zurich.\\
$^{2}$Technische Universit\"at Berlin, Einstein Center Digital Future and SCIoI Excellence Cluster, Germany.
\vspace{1ex}
}

\begin{document}
\twocolumn[{%
\renewcommand\twocolumn[1][]{#1}%
\maketitle
\centering
\vspace{-0.9cm}
\includegraphics[trim=10.0 0.5 0 0, clip, width=0.9\linewidth]{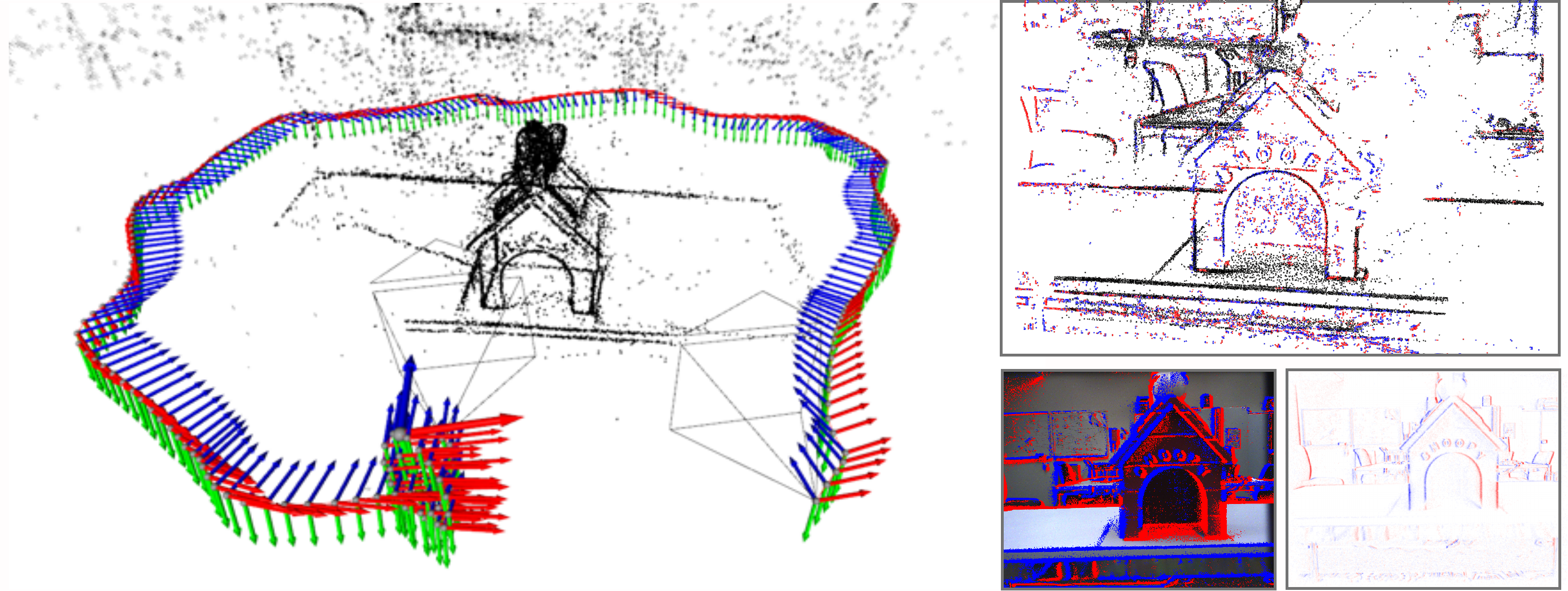}
\vspace{-1ex}
\captionof{figure}{\label{fig:eyecatcher}Camera trajectory and estimated 3D map (left). The top-right
inset shows the sliding window map (grayscale points) with the current keyframe
map (pseudo-colored blue-red points, according to event polarity). The
bottom-right insets show the color image (frame) with the real events (left) and
the image obtained by the event generative model (right).
\vspace{4ex}
}
}]

\begin{abstract}
We introduce EDS, a direct monocular visual odometry using events and frames.
Our algorithm leverages the event generation model to track the camera
motion in the blind time between frames. The method formulates a direct
probabilistic approach of observed brightness increments. Per-pixel
brightness increments are predicted using a sparse number of selected 3D
points and are compared to the events via the brightness increment error to estimate
camera motion. The method recovers a semi-dense 3D map using photometric
bundle adjustment. EDS is the first method to perform 6-DOF VO using events
and frames with a direct approach. By design it overcomes the problem of
changing appearance in indirect methods. Our results outperform all previous
event-based odometry solutions. We also show that, for a target error performance,
EDS can work at lower frame rates than state-of-the-art frame-based VO solutions.
This opens the door to low-power motion-tracking
applications where frames are sparingly triggered ``on demand'' and our
method tracks the motion in between. We release code and datasets to the
public.
\end{abstract}

\vspace{-3ex}
\section*{Multimedia Material}
\noindent Code and dataset can be found at: \url{https://rpg.ifi.uzh.ch/eds}

\hyphenation{projected}

\section{Introduction}
\label{sec:intro}

Visual Odometry (VO) is a paramount tool in computer vision, robotics and, any
application that requires spatial
reasoning~\cite{Cadena16tro,Davison18arxiv,Loquercio2021Science}.  In recent
years, considerable progress has been made on this topic
\cite{MurArtal15tro,Cadena16tro,Engel17pami,Forster17troSVO}.  However, VO systems are limited by the
capabilities of their physical devices (sensors, processors, and power).  
Some of these limitations (e.g., motion blur, dynamic range) can been tackled with novel and/or more robust sensors, such as event cameras.

Event cameras~\cite{Lichtsteiner08ssc,Posch10isscc, Posch14ieee} are bio-inspired sensors that work radically different from
traditional cameras. Instead of capturing brightness images at a fixed rate,
they measure asynchronous, per-pixel brightness changes, called
``events''.\footnote{See an illustrative animation:
\url{https://youtu.be/LauQ6LWTkxM?t=30}} This principle of operation endows
event cameras with outstanding properties, such as low
latency, high temporal resolution (in the order of \si{\micro\second}) and low
power (milliwatts instead of watts). The large potential of event cameras to
tackle VO and related problems in challenging scenarios has been investigated
in~\cite{Censi14icra,Kim14bmvc,Mueggler14iros,Kim16eccv,Kueng16iros,Zhu17cvpr,Gallego17pami,Rebecq17ral,Rosinol18ral,Gallego18cvpr,Chamorro20bmvc,Kim21ral}.
We refer to a recent comprehensive survey paper for further details~\cite{Gallego20pami}.

Event-based VO is a challenging problem that has been
addressed step-by-step in scenarios with increasing complexity.
Two fundamental \emph{challenges} arise when working with event cameras: noise
(caused by timestamp jitter, pixel manufacturing mismatch or non-linear circuity
effects) and data association~\cite{Gallego20pami,Gehrig19ijcv}, i.e.,
establishing correspondences between events to identify which events are
triggered by the same scene point. This is so because each event carries little
information and the temporal edge patterns conveyed by the events depend on
motion.  These two issues make event-based keypoints used by indirect methods
difficult to detect and track with sufficient stability; for this reason,
grayscale frames~\cite{Kueng16iros} or motion compensation and inertial sensor
fusion~\cite{Rosinol18ral} have been used to mitigate their effect.

Event-based methods might be categorized according to whether they exploit the
Event Generation Model (EGM)~\cite{Cook11ijcnn,Kim16eccv} or
not~\cite{Kueng16iros,Rebecq17ral,Zhu19cvpr,Zhou21tro}.  The EGM states how
events are created when a predefined contrast threshold is
reached~\cite{Lichtsteiner08ssc,Gallego20pami}.  It is a photometric
relationship between brightness ``changes'' (i.e., events) and ``absolute''
brightness.  Experiments on event-based feature tracking~\cite{Gehrig19ijcv}
have shown that methods exploiting the EGM achieve higher accuracy than those
that do not.  However such a comparison is yet to be performed for 6-DOF camera
tracking (i.e., ego-motion estimation). Current
solutions~\cite{Kim16eccv,Rebecq17ral} are prone to lose tracking, either
because the convergence of the estimated 3D map is slow~\cite{Kim16eccv} or
because the edge-patterns change quickly from one packet of events to the next
one~\cite{Rebecq17ral}.  That is, there is no long-term appearance, like the
grayscale frames in~\cite{Gehrig19ijcv}, to latch onto and improve tracking
robustness, which we intend to do.

We propose to tackle the event-based VO problem by understanding and overcoming
the shortcomings of previous methods.  To the best of our knowledge, this work
is the first monocular method to perform 6-DOF VO using events and frames with a
direct approach. Our contribution lies in the front-end, fusing the information
from events and frames tightly using the EGM (as opposed to
previous works that loosely coupled them \cite{Rosinol18ral}). 
Estimated poses, points, and selected
frames are fed into a sliding-window photometric bundle adjustment (PBA)
back-end, which minimizes brightness errors. This is the first time that PBA is used
in the context of event-camera VO.  Internally, we adopt a key-frame-based
approach, recovering inverse depth at pixels with high gradient as in DSO~\cite{Engel17pami}. 
However, our method allows tracking the camera motion in
the blind time between frames using only events. This opens the door for
frames to be triggered~\emph{sparingly}, i.e., ``on demand'', thus potentially
saving energy in the system, which is a desirable feature in AR/VR applications
and in platforms with a limited power budget. 
Our contributions are summarized as follows:
\begin{itemize}
    \item The first formulation of a monocular 6-DOF visual odometry combining
        events and grayscale frames in a direct approach with photometric bundle
        adjustment.

    \item Camera motion tracking using a sparse set of pixels, minimizing the
        normalized brightness change of projected (inverse depth) points.

    \item A compelling evaluation on publicly available datasets outperforming
        previous solutions, and a sensitivity study to gain insights about our method.

    \item A new dataset with high quality events, color frames, and IMU data to foster research in monocular VO.
\end{itemize}

\hyphenation{pixels}

\section{Related Work}
\label{sec:related}

\begin{table}[tb!]
\centering
\begin{adjustbox}{max width=1.0\linewidth}
\setlength{\tabcolsep}{3.5pt}
\begin{tabular}{lccccl}
 & \textbf{Events} & \textbf{Frames} & \textbf{D/I} & \textbf{EGM} & \textbf{Remarks}\\
\midrule
Kim et al.~\cite{Kim16eccv} & \yesmark & \nomark & D & \yesmark & Three parallel EKFs\\ %
Rebecq et al.~\cite{Rebecq17ral} & \yesmark & \nomark & D & \nomark& Parallel tracking and mapping\\
Kueng et al.~\cite{Kueng16iros} & \yesmark & \yesmark & I & \nomark & Tracking event features for VO\\ %
Rosinol et al.~\cite{Rosinol18ral} & \yesmark & \yesmark & I & \nomark & Loosely coupled front-end\\
\textbf{This work} & \yesmark & \yesmark & D & \yesmark & Tightly coupled front-end\\
\bottomrule
\end{tabular}
\end{adjustbox}
\vspace{-1ex}
\caption{\emph{Comparison of event-based monocular 6-DOF VO/SLAM methods}. 
The columns indicate the type of input (events and/or grayscale frames), 
the type of method (\textbf{D}irect or \textbf{I}ndirect),
and whether the method exploits the event generation model (EGM).
\label{tab:eslam:methods}
}
\vspace{-1ex}
\end{table}

Event-based VO methods might be \emph{direct} or \emph{indirect} depending on whether
they process raw pixel information \emph{directly} or \emph{indirectly} via some
intermediate representation.  Indirect methods~\cite{Kueng16iros,Zhu17cvpr},
like frame-based approaches,
extract keypoints~\cite{Mueggler17bmvc,Alzugaray18ral} from the input (event)
data in the front-end before passing them to the back-end. Direct
methods~\cite{Kim16eccv,Rebecq17ral} attempt to directly process all events
available. Since events correspond to per-pixel brightness changes, they
naturally convey the information about the motion of the scene edges (assuming
constant illumination).
Early event-based VO works \cite{Cook11ijcnn,Weikersdorfer13icvs,Kim14bmvc}
tackled simple camera motions, such as \mbox{3-DOF} (planar or rotational), and
hence did not account for depth.  The most general case of a freely moving
camera (\mbox{6-DOF}) has been tackled only
recently~\cite{Kueng16iros,Kim16eccv,Rebecq17ral}.  In terms of scene texture,
high contrast and/or structured scenes have been addressed before focusing on
more difficult, natural 3D scenes.  Event-based VO is still in its infancy, with
the majority of works addressing only camera
tracking \cite{Weikersdorfer12robio,Mueggler14iros,Mueggler15rss,Gallego17pami,Bryner19icra,Chamorro20bmvc}
because of its simplicity.
\Cref{tab:eslam:methods} summarizes the related work on event-based monocular
\mbox{6-DOF} tracking and mapping.

\textbf{Why direct methods with events?}
Feature-based methods work well for standard cameras~\cite{MurArtal17tro}, where
features are mature and easy to detect and track due to small intensity noise.
However they are surpassed in accuracy by direct methods~\cite{Engel17pami}
(which use all data available, even pixels that do not comply with the
definition of a feature).  In event cameras, features are not easy to detect and
track because the event ``appearance'' depends on motion (and
texture)~\cite{Gallego20pami}.  There are many event-based feature detectors
and/or
trackers~\cite{Vasco16iros,Mueggler17bmvc,Alzugaray18ral,Chiberre21cvprw}, but
their application to enable VO is scarce~\cite{Kueng16iros} because they are not
as accurate and stable as needed.  On the other hand, ($i$) events are triggered
by moving edges (which are semi-dense on the image plane), and ($ii$) semi-dense
methods are state of the art for event-based 3-DOF VO
\cite{Weikersdorfer13icvs,Gallego17ral,Reinbacher17iccp,Kim21ral}.  All these
ideas suggest that direct methods are the natural fit for event cameras and
should also work well for 6-DOF motion estimation, as hinted by early
works~\cite{Rebecq17ral,Kim16eccv}.

\textbf{Combining events and frames}.
Events and intensity frames are complementary sources of visual
information~\cite{Scheerlinck18accv,Ni12tro}.  Combining them has proven useful
to improve accuracy and/or robustness in several applications, such as feature
tracking~\cite{Gehrig19ijcv}, ego-motion
estimation~\cite{Bryner19icra,Rosinol18ral}, depth
prediction~\cite{Gehrig21ral}, video
reconstruction~\cite{Scheerlinck18accv,Pan20pami} and video
frame interpolation~\cite{Tulyakov21cvpr}.  We also seek to
get the best of both visual sensors for VO-related tasks, as implied
by~\cite{Bryner19icra,Rosinol18ral}.  In contrast to~\cite{Rosinol18ral}, which
treats events and frames as unrelated visual sources (i.e., no effort is made in
the front-end to fuse the same feature seen by both sensors), we fuse events and
frames in a principled way in the VO front-end by exploiting the
EGM~\cite{Gallego15arxiv,Gehrig19ijcv}.  Hence, our approach is closer
to~\cite{Bryner19icra}, but the map is not externally given and errors are computed on
a sparse set of pixels in the keyframes.

In short, as \Cref{tab:eslam:methods} shows, there is a gap in the
characteristics of the approaches used to tackle the monocular 6-DOF VO problem, and
this work fills that gap: fusing events and frames in the front-end via a direct
method to exploit the EGM. Our method includes a back-end with a photometric bundle
adjustment~\cite{Engel17pami} adapted to the front-end, differing from
the feature-based back-end in~\cite{Rosinol18ral}. To the best of our knowledge, 
this path has not yet been explored.

\hyphenation{pixelwise}

\section{Direct Odometry with Events and Frames}
\label{sec:method}

\begin{figure*}[tb]
    \centering
    \includegraphics[width=0.99\linewidth]{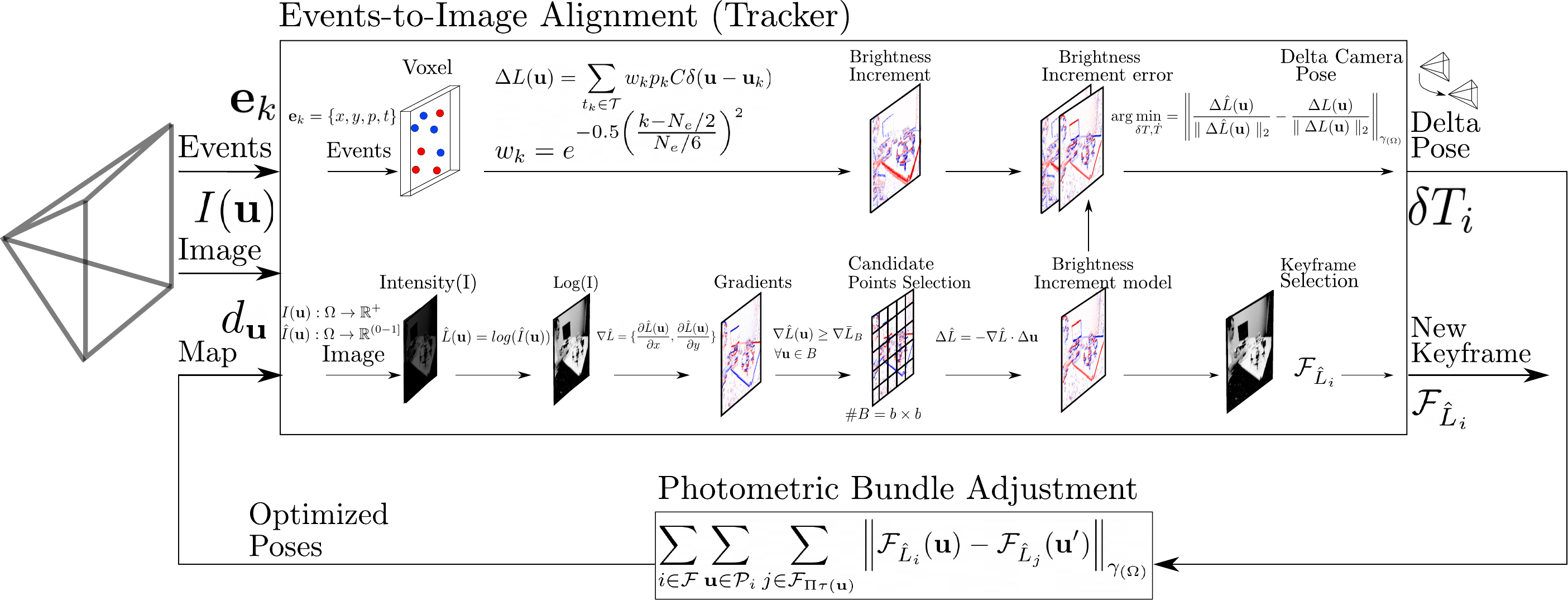}
    \caption{Block diagram of the proposed event-based direct odometry approach.
    Events and frames acquired by a camera, such as the DAVIS346~\cite{Taverni18tcsii}, are fed to the front-end, 
    where they are fused using the event generation model (EGM). 
    The front-end selects sparse points on scene edges (i.e., events) with respect to the keyframes. 
    As the camera moves, it generates events, and the camera pose is estimated with respect to the last keyframe. Poses and keyframes are passed to the back-end, which performs non-linear refinement of the poses and depth estimates via photometric bundle adjustment. 
    These are later fed back to the front-end to sustain the good performance of the VO system.
    Events are colored in blue/red according to polarity $p_k$, indicating positive/negative brightness increments.
    }
    \label{fig:method}
\end{figure*}

This section describes our method, which is summarized in the block diagram of
\cref{fig:method}. First, we review how an event camera works and the EGM
(\cref{sec:method:generationmodel}). Then we describe the system's front-end
(\cref{sec:method:frontend}), back-end (\cref{sec:method:backend}) and
initialization (\cref{sec:method:bootstrapping}).

\subsection{Event Generation Model (EGM)}
\label{sec:method:generationmodel}

\textbf{How an Event Camera Works}.
Each pixel of an event camera produces an event $e_k\doteq (\bu_k,t_k,\pol_k)$
whenever it detects that the logarithmic brightness
$\Lum$ at that pixel, $\bu_k = (x_k,y_k)^\top$, changes by a specified amount $C$ (called contrast
sensitivity)~\cite{Lichtsteiner08ssc}:
\begin{equation}
\label{eq:generativeEventCondition}
\Delta \Lum(\bu_k,t_k) \doteq \Lum(\bu_k,t_k) - \Lum(\bu_k, t_k-\Delta t_k) = \pol_k \, C,
\end{equation} 
where the polarity $\pol_k \in \{+1,-1\}$ is the sign of the brightness change,
and $\Delta t_k$ is the time elapsed since the last event at the same pixel.
Event timestamps $t_k$ have \si{\micro\second} resolution. A single pixel has
its own sampling rate (which depends on the visual input) and produces events
proportionally to the amount of scene motion.  An event camera does not output
images at a constant rate, but rather a stream of asynchronous events in
space-time.

\textbf{Linearized Event Generation Model}. 
Collecting pixelwise the polarities of a set of events $\cE \doteq
\{e_k\}_{k=1}^{\numEvents}$ in the time interval $\cT \doteq
\{t_k\}_{k=1}^{\numEvents}$ produces a brightness increment image:
\begin{equation}
\label{eq:brightnessIncrementEvents}
\Delta \Lum(\bu) =  \sum_{t_k\in \cT} \pol_k C\, \delta(\bu-\bu_k),
\end{equation}
where the Kronecker $\delta$ selects the appropriate pixel.  If the number of
events $\numEvents$ spans a small delta time $ \Delta t = t_{\numEvents} - t_1$,
the increment~\eqref{eq:generativeEventCondition} can be approximated using Taylor's
expansion.  Further substituting the brightness constancy assumption gives that
$\Delta \Lum$ is caused by brightness gradients $\nabla\Lum$
moving with velocity $\velflow$ on the image plane~\cite{Gehrig19ijcv,Gallego15arxiv}:
\begin{equation}
\label{eq:brightnessIncrementGrad}
\Delta \Lum(\bu) \approx - \nabla \Lum(\bu) \cdot \Delta \bu = - \nabla \Lum(\bu) \cdot \velflow(\bu) \Delta t.
\end{equation}

\subsection{Front-End}
\label{sec:method:frontend}

In the VO front-end (\cref{fig:method}) we use
\eqref{eq:brightnessIncrementEvents} to create pseudo-measurements from the
events (top branch of \cref{fig:method}), and use the right hand side of
\eqref{eq:brightnessIncrementGrad} to predict those from the frame $\hat{\Lum}$
and the current state of the VO system (middle branch of \cref{fig:method}).
Our goal is, roughly speaking, to estimate the VO state that best predicts the
measurements (\cref{fig:event_model_anim}).
This strategy is described in the upcoming subsections.

\textbf{Event Weighting}. 
As pointed out in~\cite{Bryner19icra}, there is a trade-off in the selection of
$\numEvents$ to build~\eqref{eq:brightnessIncrementEvents}: a small $\numEvents$
does not yield sufficient SNR or evidence of existing edge motion, whereas a
large $\numEvents$ produces accumulation blur and breaks the assumption about
events being triggered by the camera at a single location.
To tackle this issue, we select $\numEvents$ large to have enough SNR and we
modify~\eqref{eq:brightnessIncrementEvents} to accumulate weighted polarities
$w_k p_k \leftarrow p_k$.  We use Gaussian weights $w_k$ in time index $k$ (top
branch in \cref{fig:method}).  The weights emphasize the central part of the
window of events $\cE$, hence producing thinner edges (less accumulation blur)
than in the unweighted case ($w_k=1$).

\textbf{Events Prediction using Frames}.
\label{sec:eventpredictions}
The middle branch of \cref{fig:method} computes the right hand side of
\eqref{eq:brightnessIncrementGrad} and selects only pixels at scene contours for event prediction. 
A keyframe in the middle branch of \cref{fig:method} comprises a brightness frame and a (semi-dense) inverse depth map (\cref{sec:method:mapping}).

The spatial gradient~\eqref{eq:brightnessIncrementGrad} of the
logarithmic normalized intensity of the keyframe $\hat{\Lum}$ is computed using the Sobel operator.
The image-point velocity $\velflow$ 
in~\eqref{eq:brightnessIncrementGrad} is purely geometric, given in terms
of the camera pose $\pose$, the camera's linear and angular velocities $\vel \doteq
(\linvel^\top,\angvel^\top)^\top$, and the depth $\depthu \doteq Z(\bu)$ of the 3D point with
respect to the camera~\cite{Corke17book}:%
\begin{equation}
\label{eq:MotionField}
\velflow (\bu) = \mat{J}(\bu, \depthu) \,\vel,
\end{equation}
where $\mat{J}(\bu, \depthu)$ is the $2\times 6$ feature sensitivity matrix
\begin{equation}
\label{eq:feature_sensitivity_matrix}
\mat{J} (\bu,\depthu) =
\scalemath{0.85}{
\begin{bmatrix}
    \frac{-1}{\depthu} & 0    & \frac{u_x}{\depthu} & u_x u_y & -(1+u_x^2) &  u_y \\[1ex]
        0    & \frac{-1}{\depthu} & \frac{u_y}{\depthu} & 1+u_y^2 & -u_x u_y   & -u_x
\end{bmatrix}
}.
\end{equation}

Inserting~\eqref{eq:MotionField} in~\eqref{eq:brightnessIncrementGrad} gives the predicted brightness change%
\begin{equation}
\label{eq:brightnessIncrementGrad2}
\Delta \hat{\Lum}(\bu) \approx - \nabla \hat{\Lum}(\bu) \cdot \mat{J}(\bu,\depthu) \vel \Delta t.
\end{equation}
The pose $\pose$ and velocity $\vel$ are global quantities shared by all image
pixels $\bu$.  The keyframe brightness $\hat{L}$ and the delta time $\Delta t$
(given by the event timestamps) are known.  As noted in the block diagram of
\cref{fig:method}, the depth $\depthu$ is an input to the front-end,
given by the back-end. Initialization of the depth estimates is described in
\cref{sec:method:bootstrapping}.

\textbf{Candidate Points Selection}. 
Only the most informative pixels of the keyframes are used.  This focuses
computational resources while maintaining accuracy.  Specifically, we select
pixels with a sufficiently strong gradient (i.e., contours).
To have a good distribution of the selected pixels over the image plane, we
follow a tiling approach, dividing the image into rectangular tiles (e.g.,
$11\times 11$ tiles) and selecting a percentage of the pixels with the largest
brightness gradient on each tile (typically 10-15\% of the image pixels). 
Note that at a keyframe the pixels with strongest gradients overlap with the
pixels where events are triggered (since events are due to moving edges).  As
the camera moves, these two sets of pixels begin to depart; however, it is
through the estimation of the camera motion (\cref{sec:method:tracking}) and the
scene depth (\cref{sec:method:mapping}) that we keep them aligned, thus
maintaining a correspondence between them.

\subsubsection{Camera Tracking}
\label{sec:method:tracking}

Camera tracking (illustrated on the top right of the front-end block in
\cref{fig:method}) is performed with respect to the last keyframe.  We split the
event stream into packets (i.e., temporal windows), and create event
frames~\eqref{eq:brightnessIncrementEvents} using the Gaussian weighting
mentioned above.  We cast the camera tracking problem as a joint optimization
over the camera motion parameters (6-DOF pose and its velocity):
\begin{equation}
\label{eq:poseestim:argmin}
(\deltapose^{\ast}, \vel^{\ast}) = \arg\min_{\deltapose, \vel} \left\| \frac{\Delta \hat{\Lum}}{\|\Delta
    \hat{\Lum}\|_2} - \frac{\Delta \Lum}{\|\Delta \Lum\|_2} \right\|_\gamma.
\end{equation}
The error is the Huber norm $\gamma$ of the difference between normalized brightness
increments ($\Delta \Lum$ from the events and $\Delta \hat{\Lum}$ from the
keyframe).  Norms are computed over a sparse set: the above-mentioned
selected pixels in the keyframe.  Hence, the increments due to the events are
transferred to the image plane of the keyframe: ($i$) first, we find the
location $\bu_e$ on the event image plane corresponding to a selected keyframe
pixel $\bu_f$:
\begin{equation}
\label{eq:xe_from_xf}
\bu_e = \pi\bigl(T_{e,f} \, \pi^{-1}\bigl(\bu_f,d_{\bu_f}\bigr) \bigr),
\end{equation}
where $\pi^{-1}:\Real^2\times\Real \to\Real^3$ back-projects a keyframe pixel, 
$T_{e,f} \in SE(3)$ changes coordinates to the current event camera pose,
and $\pi:\Real^3\to\Real^2$ projects the point onto the event camera;
and ($ii$) we compute $\Delta \Lum(\bu_e)$ by cubic interpolation of the brightness increment~\eqref{eq:brightnessIncrementEvents}.
The normalized values of $\Delta \Lum(\bu_e)$ and $\Delta \hat{\Lum}(\bu_f)$ are compared in~\eqref{eq:poseestim:argmin}.

The pose $\pose \in SE(3)$ is parametrized using a 3-vector and a quaternion.
The velocity $\vel$ is parametrized using a 6-vector.  To
minimize~\eqref{eq:poseestim:argmin}, we use the Ceres
solver\cite{ceres-solver}, which performs local Lie group parametrization of $T$.

Compared to the camera tracker~\cite{Bryner19icra}, which projects a global
photometric 3D map onto the current event camera location, we compute the
error~\eqref{eq:poseestim:argmin}: ($i$) on the keyframe (local depth maps),
rather than on the event frame (because we lack dense optical flow from a
global 3D map to use in the EGM), ($ii$) only in a sparse group of pixels
(as opposed to the entire image plane) conforming a local semi-dense map.

The output of the front-end is the current keyframe (brightness image and
inverse depth map) and the estimated camera motion (relative to the keyframe) of
each event packet.  These are passed to the back-end for further non-linear
refinement (\cref{sec:method:backend}).

\textbf{Keyframe Selection}. 
A keyframe is created when one of two conditions is met: ($i$) the number of
selected points decreases by 20-30\% (because they fall out of the field of view (FOV)).
($ii$) the relative rotation of the event camera with respect to the keyframe exceeds a given threshold.

\subsubsection{Mapping (Depth Estimation)}
\label{sec:method:mapping}
As a new keyframe is created, the inverse depth estimates from past keyframes
are used to populate those of the new keyframe.  Likewise, the set of selected
pixels is transferred to the new keyframe (akin to~\eqref{eq:xe_from_xf}).  The
inverse depth values at the remaining pixels are initialized using nearest
neighbors with a $k$-d tree.  This is simple and effective.  In our VO system,
inverse depth estimates are refined in the back-end.

\subsection{Back-End (Non-linear Refinement)}
\label{sec:method:backend}

The back-end (bottom branch in \cref{fig:method}) performs non-linear refinement of the camera poses and the 3D structure
via photometric bundle adjustment (PBA).
It minimizes the objective function
\begin{equation}
\sum_{i \in \mathcal{F}} \sum_{\bu \in \mathcal{P}_i} \sum_{j \in
    \mathcal{F}_{\Pi\boldsymbol{\tau}(\mathbf{u})}}  \left \|
    \mathcal{F}_{\hat{L}_i}(\bu) - \mathcal{F}_{\hat{L}_j}(\mathbf{u^\prime}) \right \|_{\gamma_{(\Omega)}},
\end{equation}
where $i\in \cF$ runs over all keyframes $\cF$, $\bu$ runs over all selected
pixels $\mathcal{P}_i$ in keyframe $i$, $j$ runs over all keyframes in which
point $\bu$ is visible, and $\bu'$ is the corresponding point to $\bu$ on the
$j$-th keyframe.  We confer robustness by using the Huber norm $\gamma$, which
deemphasizes bad correspondences, and by discarding outliers (based on
obnoxiously large errors).  Errors are measured around each image point using an
8-pixel patch and assuming the same depth estimate for all pixels in the patch
(residual pattern \#8 in~\cite{Engel17pami}).

$N_k$ keyframes are kept in a sliding window estimator (we use 7, as
in~\cite{Engel17pami}, since it is a good accuracy-efficiency compromise).  On
average, around 2000--8000 points are used in the back-end.
We have implemented a PBA using Ceres~\cite{ceres-solver} with automatic
differentiation.
We have also combined our front-end with DSO's PBA~\cite{Engel17pami}, since our design is modular.

\subsection{Bootstrapping}
\label{sec:method:bootstrapping}
To initialize the system, we may try three methods on the frames: ($i$) classical
multi-view geometry, ($ii$) learning-based monocular depth prediction or ($ii$) DSO's coarse initializer. 
The first approach uses the 8-point algorithm~\cite{Hartley03book} for the first frames until the sliding window is full (we skip frames to increase
parallax). Once initialization is successful, the frames become keyframes,
we select points with large gradient, and perform one PBA step to refine the map.
If the above fails (i.e.: planar scenes), we use single image depth prediction via
MiDAS~\cite{Ranftl20pami}. 
Ultimately, DSO's coarse initializer works the best to initialize. 
The magnitude of the predicted depth is arbitrary, but this is known in
monocular VO since scale is unobservable (a Gauge freedom \cite{Zhang18ral}).

\hyphenation{several}
\hyphenation{evaluation}
\hyphenation{methods}

\section{Experiments}
\label{sec:experim}

We now evaluate the designed monocular VO method.  First, we present the
datasets and metrics used for the evaluation (\cref{sec:experim:datasets}).
Second, we introduce the baseline methods (\cref{sec:experim:baselines}).
Third, we evaluate the performance of the method, comparing with the state of
the art (\cref{sec:experim:tracking}). Finally, we analyze the sensitivity of
the results to various perturbations and limitations in
\cref{sec:experim:study}.

\subsection{Datasets and Evaluation Metric}
\label{sec:experim:datasets}
We test on sequences from the standard dataset\footnote{RPG Stereo DAVIS:
\url{https://rpg.ifi.uzh.ch/ECCV18_stereo_davis.html}}~\cite{Zhou18eccv} to assess the
performance of the proposed VO method. Data provided by \cite{Zhou18eccv} was
collected with a hand-held stereo DAVIS240C in an indoor environment, and ground
truth poses were given by a motion capture system with sub-millimeter accuracy.
See the supplementary material for collected sequences with our custom-made beamsplitter.
To assess the performance of the full VO method, we report ego-motion estimation
results using standard metrics: 
absolute trajectory error (ATE) and rotation RMSE~\cite{Sturm12iros}.
We use the toolbox from~\cite{Zhang18iros} to evaluate the poses given by
different visual odometry solutions. Monocular methods are tested using data
from the left camera only.  The tracking and mapping parts of the VO methods are
connected, so depth estimation errors are subsumed in the accuracy of the
estimated camera trajectory, in a tightly coupled manner. 

\begin{figure}[t]
    \centering
    \ifcompileanimated
      \animategraphics[width=0.49\linewidth]{10}{imgs/events/frame_00000008}{00}{49}
      \animategraphics[width=0.49\linewidth]{10}{imgs/model/frame_00000008}{00}{49}
    \else 
      \includegraphics[width=0.49\linewidth]{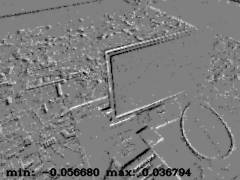}
      \includegraphics[width=0.49\linewidth]{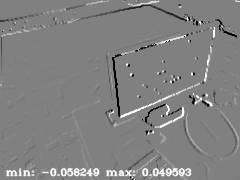}
    \fi
    \vspace{-1ex}
    \caption{Event frames \eqref{eq:brightnessIncrementEvents} (left) 
    and EGM frames \eqref{eq:brightnessIncrementGrad2} (right) 
    for the \rpgmonitor{} sequence.
    Here, positive brightness changes are displayed in white, and negative ones in black.
    Gray color means there is no brightness change.
    This figure contains animations that can be viewed in \emph{Acrobat Reader}.}
    \label{fig:event_model_anim}
\end{figure}

\def\figWidth{0.192\linewidth}
\begin{figure*}[t]
	\centering
    {\small
    \setlength{\tabcolsep}{2pt}
	\begin{tabular}{
	>{\centering\arraybackslash}m{0.4cm}
	>{\centering\arraybackslash}m{\figWidth} 
	>{\centering\arraybackslash}m{\figWidth}
	>{\centering\arraybackslash}m{\figWidth}
	>{\centering\arraybackslash}m{\figWidth}}
		& 
		\rpgbin{} & 
		\rpgboxes{} & 
		\rpgdesk{} & 
		\rpgmonitor{}\\[.5ex]

        \rotatebox{90}{\makecell{\emph{EVO} \cite{Rebecq17ral}}}&
		\frame{\includegraphics[width=\linewidth]{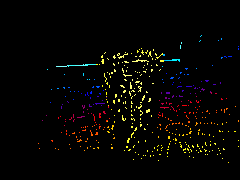}}&
		\frame{\includegraphics[width=\linewidth]{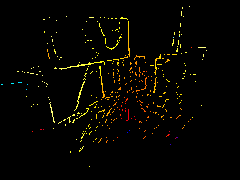}}&
		\frame{\includegraphics[width=\linewidth]{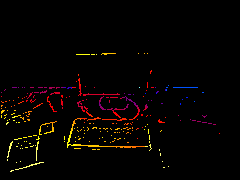}}&
		\frame{\includegraphics[width=\linewidth]{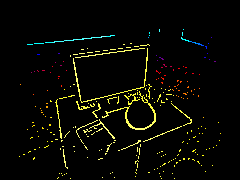}}
		\\
	
        \rotatebox{90}{\makecell{\emph{DSO} \cite{Engel17pami}}}&
		\frame{\includegraphics[width=\linewidth]{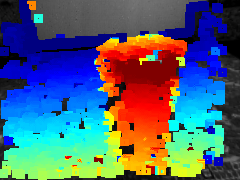}}&
		\frame{\includegraphics[width=\linewidth]{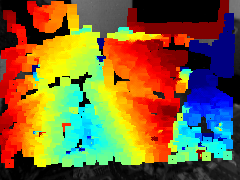}}&
		\frame{\includegraphics[width=\linewidth]{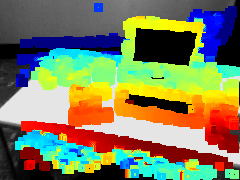}}&
		\frame{\includegraphics[width=\linewidth]{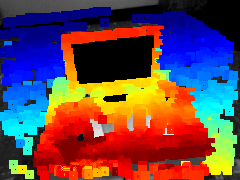}}
		\\
	
        \rotatebox{90}{\makecell{\emph{EDS (Ours)}}}&
		\frame{\includegraphics[width=\linewidth]{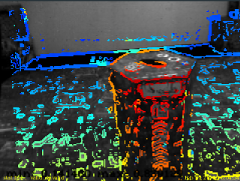}}&
		\frame{\includegraphics[width=\linewidth]{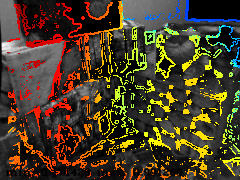}}&
		\frame{\includegraphics[width=\linewidth]{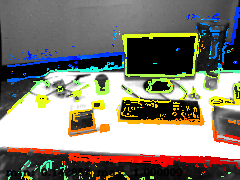}}&
		\frame{\includegraphics[width=\linewidth]{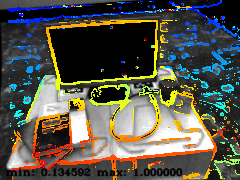}}
		\\
	    
        \rotatebox{90}{\makecell{\emph{EDS point cloud}}}&	
		\frame{\includegraphics[width=\linewidth]{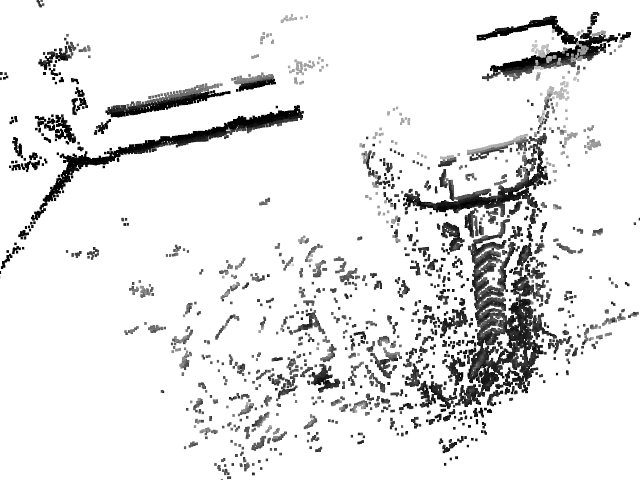}}&
		\frame{\includegraphics[width=\linewidth]{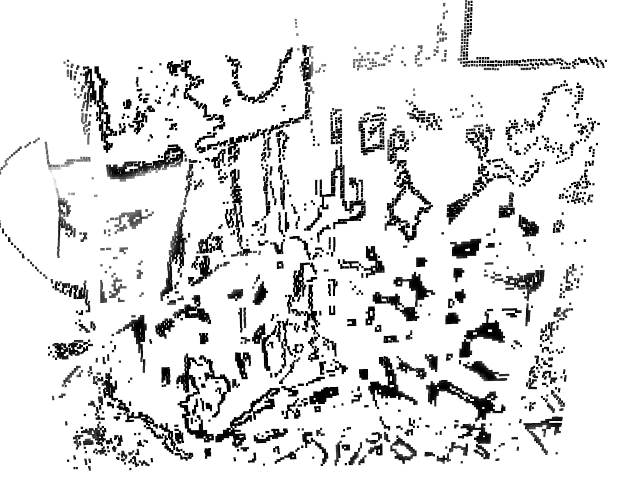}}&
		\frame{\includegraphics[width=\linewidth]{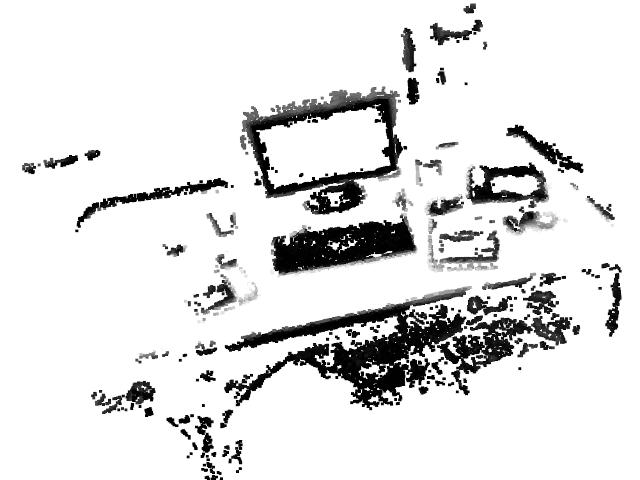}}&
		\frame{\includegraphics[width=\linewidth]{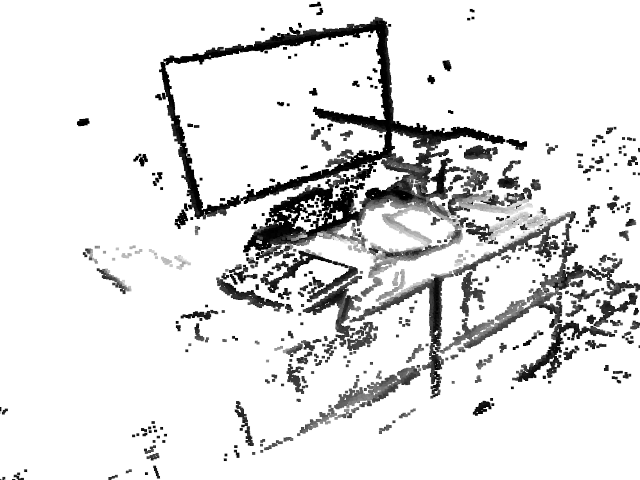}}
		\\
	\end{tabular}
	}
	\vspace{-1ex}
    \caption{Qualitative comparison on four test sequences from~\cite{Zhou18eccv}. 
    The first three rows depict pseudo-colored inverse depth maps for each method. 
    EVO's color code is yellow-near blue-far, while DSO and EDS colors are red-near blue-far. 
    The depth range is \SIrange{1}{7}{\meter} in all sequences. 
    The 3D point cloud reconstructed by EDS is shown in the last row, with grayscale values from the keyframe.}
	\label{fig:results_davis}
	\vspace{-1ex}
\end{figure*}

\subsection{Baseline Methods}
\label{sec:experim:baselines}
We compare with other methods in~\cref{tab:eslam:methods}, with stereo
event-based methods (for reference) and with event and frame-based
methods~\cite{Engel17pami,MurArtal17tro}. The event-based monocular methods
used for comparison are EVO and USLAM.

$\bullet$ EVO~\cite{Rebecq17ral} is a semi-dense VO approach that builds maps
using event-based space sweeping~\cite{Rebecq18ijcv} 
and tracks the camera motion by edge-map alignment, creating binary images of
1k-2k events and aligning them to the projected point cloud map. EVO does not
exploit the EGM~\eqref{eq:brightnessIncrementGrad} since it does not use frames
nor recovers image brightness.

$\bullet$ USLAM~\cite{Rosinol18ral} is an indirect monocular method that
combines events, frames and IMU measurements.  Its front-end converts events
into frames by motion compensation using the IMU's gyroscope and the median
scene depth.  Then FAST corners~\cite{Rosten06eccv} are extracted and tracked
\emph{separately} on the event frames and the grayscale frames, and are passed
to a state-of-the-art geometric feature-based back-end~\cite{Leutenegger15ijrr}.
USLAM is the only method that we consider with an IMU. The IMU is tightly used
in the front-end for event frame creation, and so removing it is not possible
without breaking the (robustness of the) method.

$\bullet$ ORB-SLAM~\cite{MurArtal17tro} and DSO~\cite{Engel17pami} are the 
state-of-the-art indirect and direct frame-based monocular visual odometry methods, respectively.
We also compare with these methods, using different types of frames 
(from the DAVIS or reconstructed using E2VID~\cite{Rebecq19pami}).

$\bullet$ EDS performs event-to-image alignment in the front-end
using $20$kevents with an overlap of 50\%, generating a new
event frame and tracking the camera motion every new $10$kevents. A
visualization is shown in \cref{fig:event_model_anim}. This gives an effective
(and adaptive) event-frame rate of $\approx 60$ FPS depending on camera motion
while the standard frames are given at a fixed rate of~\SI{50}{\milli\second}
(i.e., $20$ FPS). 

\subsection{Ego-motion Estimation Results}
\label{sec:experim:tracking}

\begin{table}[t]
\centering
\begin{adjustbox}{max width=\linewidth}
\setlength{\tabcolsep}{6pt}
\begin{tabular}{llcccc}
    & & ESVO \cite{Zhou21tro} & USLAM \cite{Rosinol18ral} & EVO \cite{Rebecq17ral} & EDS (\textbf{Ours})\\
    & \textbf{Input} & E+E & E+F+I & E & E+F\\ 
\midrule
\multirow{4}{*}{\rotatebox[origin=c]{90}{{Trans. [cm]}}}
    & \rpgbin{}     & 2.8 & 7.7 & 13.2$^\ast$ & \textbf{1.1} \\
    & \rpgboxes{}   & 5.8 & 9.5 & 14.2$^\ast$ & \textbf{2.1} \\ 
    & \rpgdesk{}    & 3.2 & 9.8 & 5.2 & \textbf{1.5} \\
    & \rpgmonitor{} & 3.3 & 6.5 & 7.8 & \textbf{1.0} \\
\midrule
\multirow{4}{*}{\rotatebox[origin=c]{90}{{Rot. [\si{\deg}]}}}
    & \rpgbin{}     & 7.61 & 7.18 & 50.26$^\ast$ & \textbf{0.99} \\
    & \rpgboxes{}   & 9.46 & 8.84 & 170.36$^\ast$ & \textbf{1.83} \\ 
    & \rpgdesk{}    & 7.25 & 32.46 & 8.25 & \textbf{1.87} \\
    & \rpgmonitor{} & 2.74 & 7.01 & 7.77 & \textbf{0.60} \\
\bottomrule
\end{tabular}
\end{adjustbox}
\vspace{-1ex}
\caption{
\emph{Comparison with event-based 6-DOF VO methods} in terms of Absolute Trajectory
Error (RMS)~[\si{\centi\meter}] and Rotation Error (RMS)~[\si{\deg}]. 
Input data \cite{Zhou18eccv} may be: events (E), grayscale frames (F) or IMU (I).
EVO entries marked with~$^\ast$ indicate the method failed after completing at most 30\% of the sequence.
}
\label{tab:compare:events}
\vspace{-1ex}
\end{table}

\begin{table}[tb]
\centering
\begin{adjustbox}{max width=\linewidth}
\setlength{\tabcolsep}{3pt}
\begin{tabular}{llccccc}
    & & ORB-SLAM~\cite{MurArtal17tro} & ORB-SLAM~\cite{MurArtal17tro} & DSO~\cite{Engel17pami} & DSO$^\dagger$ & EDS (\textbf{Ours})\\
    & \textbf{Input} & F+F & F & F & F$^\dagger$ & E+F\\ 
\midrule
\multirow{4}{*}{\rotatebox[origin=c]{90}{{Trans. [cm]}}}
    & \rpgbin{}     & 0.7 & 2.4 & 1.1 & - &\textbf{1.1} \\
    & \rpgboxes{}   & 1.6 & 3.9 & \textbf{2.0} & -  &2.1 \\ 
    & \rpgdesk{}    & 1.8 & 3.8 & 10.0 & 1.6 & \textbf{1.5} \\
    & \rpgmonitor{} & 0.8 & 3.1 & \textbf{0.9} & 2.1 & 1.0\\
\midrule
\multirow{4}{*}{\rotatebox[origin=c]{90}{{Rot. [\si{\deg}]}}}
    & \rpgbin{}     & 0.58 & \textbf{0.84} & 2.12 & -  & 0.99 \\
    & \rpgboxes{}   & 4.26 & 2.39 & 2.14 & -  & \textbf{1.83} \\ 
    & \rpgdesk{}    & 2.81 & 2.52 & 63.5 & \textbf{1.80} & 1.87 \\
    & \rpgmonitor{} & 3.70 & 1.77 & \textbf{0.33} & 1.54 & 0.60\\
\bottomrule
\end{tabular}
\end{adjustbox}
\vspace{-1ex}
\caption{
\emph{
    Comparison with frame-based 6-DOF VO methods} in terms of Absolute
    Trajectory Error (RMS)~[\si{\centi\meter}] and Rotation Error
    (RMS)~[\si{\deg}].  Input data may be: events (E), grayscale frames (F) or
    reconstructed frames from events using \cite{Rebecq19pami} (F$^\dagger$).
    DSO$^\dagger$ entries with a hyphen indicate failure after initialization,
    completing less than 10\% of the sequence.  Best monocular results are in
    bold.
}
\label{tab:compare:frames}
\vspace{-2ex}
\end{table}

Tables \ref{tab:compare:events}--\ref{tab:compare:frames} and \cref{fig:results_davis} report 
quantitative and qualitative results of the comparison of our method with the 
state of the art on sequences from the dataset~\cite{Zhou18eccv}. 
\cref{fig:results_davis} shows sample estimated depth maps and 
corresponding point clouds (back-projected depth maps). 
We kept the original depth map visualization of the baseline methods (EVO and DSO).

\textbf{Comparison to event-based baselines}.
Qualitatively, \cref{fig:results_davis} shows that EDS correctly estimates depth at most contour pixels, forming semi-dense structures in the colored depth maps and point clouds. 
The recovered 3D maps have a higher level of detail than those given by EVO.
USLAM produced too sparse maps to convey any visual insight.

Quantitatively (\cref{tab:compare:events}), 
EDS outperforms all other monocular baseline methods, even without using inertial measurements 
(which are known to improve robustness and increase accuracy in VO~\cite{Forster17troOnmanifold}).
Our approach also outperforms the state-of-the-art event-only stereo method 
ESVO~\cite{Zhou21tro}, despite the fact that our method is monocular 
and hence does not exploit the spatial parallax of stereo setups. 
The tight fusion in the front-end (between frames and events) and the PBA in the back-end compensate for the lack of stereo baseline in the event data. 

\textbf{Comparison to frame-based baselines}.
Qualitatively (\cref{fig:results_davis}), DSO produces dilated depth maps for visualization.
Hence, they look more complete, but they have also more outliers than those produced by EDS.

Quantitatively (\cref{tab:compare:frames}), 
in terms of translation errors our method is consistently better than monocular ORB-SLAM, and only slightly worse than stereo ORB-SLAM (``F+F'') with bundle adjustment enabled~\cite{MurArtal17tro}, which is on average the best performing method.
We obtain similar results as DSO, being more accurate (ATE~$1.5$\si{\centi \meter}) on the \rpgdesk{} sequence due to a faster camera motion. 
We also run DSO on reconstructed frames at $60$ FPS using E2VID~\cite{Rebecq19pami}, on the same
$20$kevents as EDS. 
The results show a lower ATE in the \rpgdesk{} sequence (low texture scene) for DSO$^\dagger$. 
However it fails in the sequences with high texture (\rpgbin{} and \rpgboxes{}), where E2VID has difficulties to correctly reconstruct the frames due to the limitations of the trained model.
Our findings agree with those of EKLT (see \cite[Tab.~7]{Gehrig19ijcv} for further details).
In terms of rotation errors, EDS is in line with the baselines. 
Events help estimating camera rotation, especially during sudden motions. 
This is the reason why DSO$^\dagger$ and EDS produce the most accurate results in the \rpgdesk{} sequence.
The large rotational error for DSO in \rpgdesk{} is due to an insufficient frame rate for the given camera speed.

\textbf{Low Frame Rate Experiments}.
Increasing motion speed makes frames further apart, hence it is effectively equivalent to reducing frame rate 
(except for motion blur), and event-based odometry excels 
at high-speed motion~\cite{Gallego17pami}, where usually frame-based approaches fail. 
Therefore, next, we progressively reduce the frame rate and compare with the state-of-the-art monocular methods from~\cref{tab:compare:frames}. 

\begin{figure}[t]
    \centering
    \includegraphics[trim={.5cm 0.1cm 1.6cm 1.3cm},clip,width=.8\linewidth]{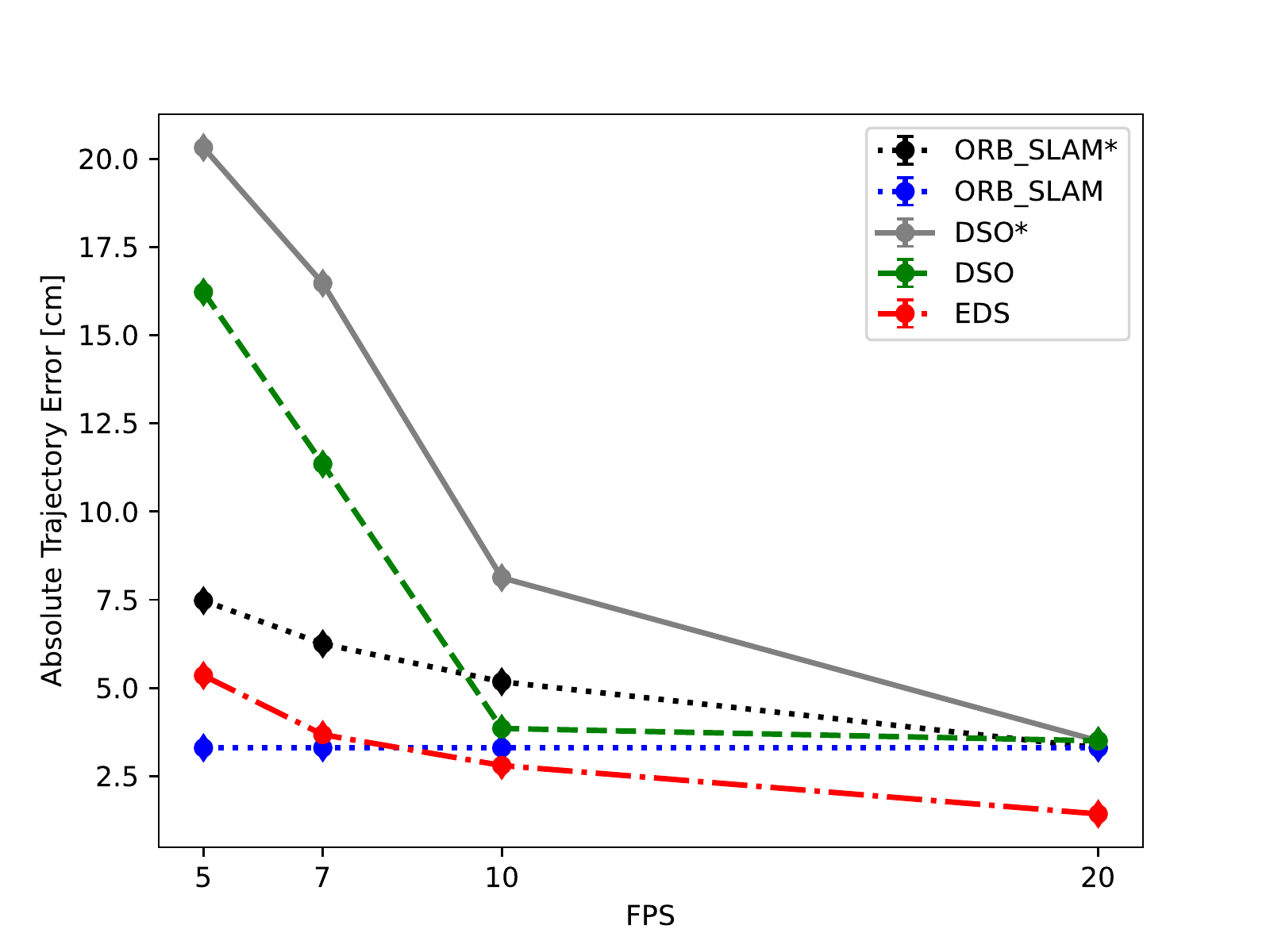}
    \vspace{-1.5ex}
    \caption{Average Absolute Trajectory Error (RMS) for ORB-SLAM~\cite{MurArtal17tro}, ORB-SLAM$^\ast$~(w/o loop closure), DSO~\cite{Engel17pami}, DSO$^\ast$~(w/o recovery-tracking) and EDS~(ours) on dataset~\cite{Zhou18eccv}.
    Errors are computed for EDS every time an event-frame is tracked (equal at any frame rate), whereas for the baseline frame-based methods they are computed only when a frame is received (i.e., according to the frame rate).
    }
    \label{fig:fps_study}
    \vspace{-1ex}
\end{figure}

We run DSO in two modes: ($i$) DSO with tracking recovery enabled using 27 different small rotations (see Sec.~3.1 in~\cite{Engel17pami})
and ($ii$) DSO with tracking recovery disabled, indicated as DSO$^\ast$. 
We also compare with ORB-SLAM (full SLAM system, for completeness) and ORB-SLAM$^\ast$ (a more fair baseline with loop closure disabled and the same number of nodes in the covisibility graph as our sliding window, i.e., 7 keyframes).
A comprehensive visualization of the results is shown in \cref{fig:fps_study}, 
where the average ATE for all sequences is depicted at the 
different frame rates (including the original frame rate from \cref{tab:compare:frames}).
The plot shows that EDS is almost agnostic to a decrease of the frame rate, while the errors significantly grow for DSO and DSO$^\ast$ as the frame rate decreases. 
DSO with tracking recovery enabled succeeds until $10$~FPS, but is not able to recover the camera pose at lower frame rates.
Compared to DSO$^\ast$, EDS does not require a tracking-recovery strategy since it is capable of continuously 
tracking at any frame rate using events, only limited by the computational cost.
The ORB-SLAM variants produce very competitive results and degrade more gracefully than DSO for low frame rates. 
Still, our method also performs better than ORB-SLAM$^\ast$.
Therefore, the EDS tracker is more robust than frame-based state-of-the-art odometry in this scenario.

\subsection{Sensitivity Study and Limitations}
\label{sec:experim:study}
Depth inaccuracies and noise in $C$ are the main subject of
study.  The brightness change in the EGM~\eqref{eq:brightnessIncrementGrad2}
highly depends on depth point estimation, which is reflected in
\eqref{eq:feature_sensitivity_matrix}.
This makes camera tracking more sensitive to accurate point triangulation than direct image alignment methods. 
Events are also susceptible to noise in $C$ due to manufacturing,
which causes a mismatch in the optical flow constraint~\eqref{eq:brightnessIncrementGrad}.
We use the simulator in \cite{Rebecq18corl} to generate synthetic sequences with
building-like scenes (e.g., atrium~\cite{Bryner19icra}) and to control the scene
and event camera parameters, in order to understand the strengths and
limitations of EDS.

\begin{figure}[t]
  \raggedleft
  {\includegraphics[trim={0.2cm 0.0cm 15.08cm 0.3cm},clip,width=\linewidth,height=3.2cm]{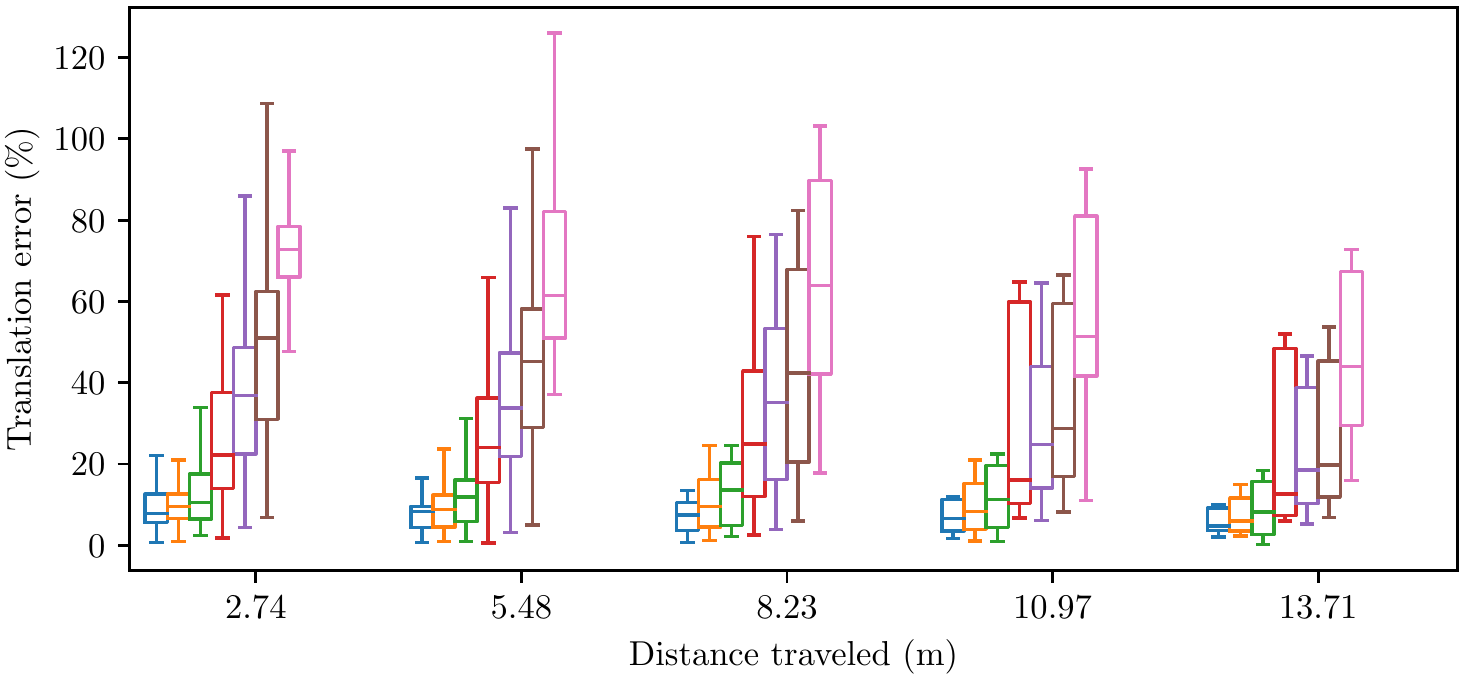}}
  {\includegraphics[trim={15.1cm 0.3cm 0.2cm 0.3cm},clip,width=\linewidth,height=3.2cm]{imgs/study/trans_and_rot_error_vs_depth.pdf}}%
  \vspace{-1ex}
  \caption{Effect of depth inaccuracies on VO (pose errors).}
  \label{fig:study:depth}
\end{figure}

\begin{figure}[t]
  \raggedleft
  \includegraphics[trim={0.2cm 0.0cm 15.25cm 0.3cm},clip,width=\linewidth,height=3.2cm]{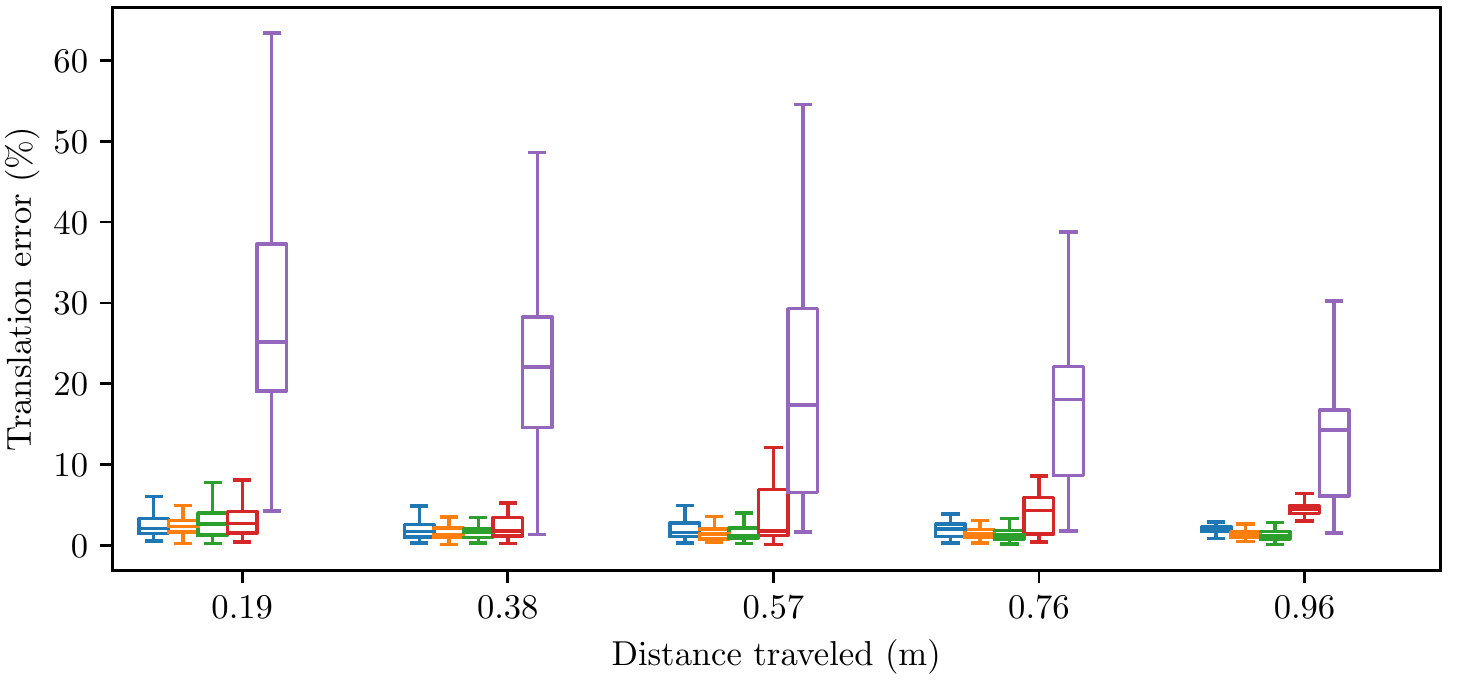}
  \includegraphics[trim={15.2cm 0.3cm 0.2cm 0.3cm},clip,width=\linewidth,height=3.2cm]{imgs/study/trans_and_rot_error_vs_noise.pdf}%
  \vspace{-1ex}
  \caption{Effect of contrast $C$ noise on VO (pose errors).}
  \label{fig:study:contrast}
\end{figure}

In one study, we perturb the ground truth 3D map with zero-mean Gaussian noise of standard deviation \SIrange{1}{50}{\percent} of the median scene depth, 
which is $9.7$~\si{\meter} in the atrium sequence~\cite{Bryner19icra}.
In another study, the contrast sensitivity is set to a mean value of $\mu_C = 0.5$ 
and is perturbed with Gaussian noise of standard deviation $\sigma_C \in [0.05, 0.25]$. 
\Cref{fig:study:depth,fig:study:contrast} show the depth and contrast errors resulting from these two sensitivity studies, respectively. 
As we observe in the box plots, the EDS tracker gracefully degrades as depth noise increases, 
while it has an abrupt degradation in terms of contrast noise, 
not being able to track the camera motion when the contrast noise has $\sigma_C > 0.15$. 
One limitation of this approach might be that events are
HDR while frames are not necessarily, which makes the EGM estimation challenging (see \cref{fig:event_model_anim}).
We refer the reader to the supplementary material for more details and experiments.

\section{Conclusion}
\label{sec:conclusion}

We have presented the first monocular direct \mbox{6-DOF} VO method using events and frames.
EDS has several innovations with respect to prior event-based methods, 
such as the tight coupling of the events and frames in the front-end
and the photometric bundle adjustment in the back-end. 
We strove to compare with multiple event- and frame-based baselines; the results showed that EDS
outperforms all event-based methods, and that it is in line with DSO at 10-20 FPS. 
However, EDS outperforms DSO and ORB-SLAM without loop closure in low frame rate scenarios, 
tracking with events accurately in between frames. 
The sensitivity study showed that EDS is robust to depth noise as well as contrast sensitivity event noise. 
We release the code and dataset to the public, and hope that this research will spark new ideas on low-power 
and robust VO by combining the best of events and frames.

\section*{Acknowledgement}
This work was supported by Huawei Zurich Research Center, the National Centre of Competence in Research (NCCR) Robotics through the Swiss National Science Foundation under grant agreement No. 51NF40\_185543, and the European Research Council (ERC) under grant agreement No. 864042 (AGILEFLIGHT).

\title{\MYTITLE\\\vspace{4ex}---Supplementary Material---\vspace{-3ex}}
\maketitle
\begin{figure*}[!ht]
\centering
    \includegraphics[width=0.33\linewidth, height=4cm, trim=0.0 0.0 0.0 0.0, clip]{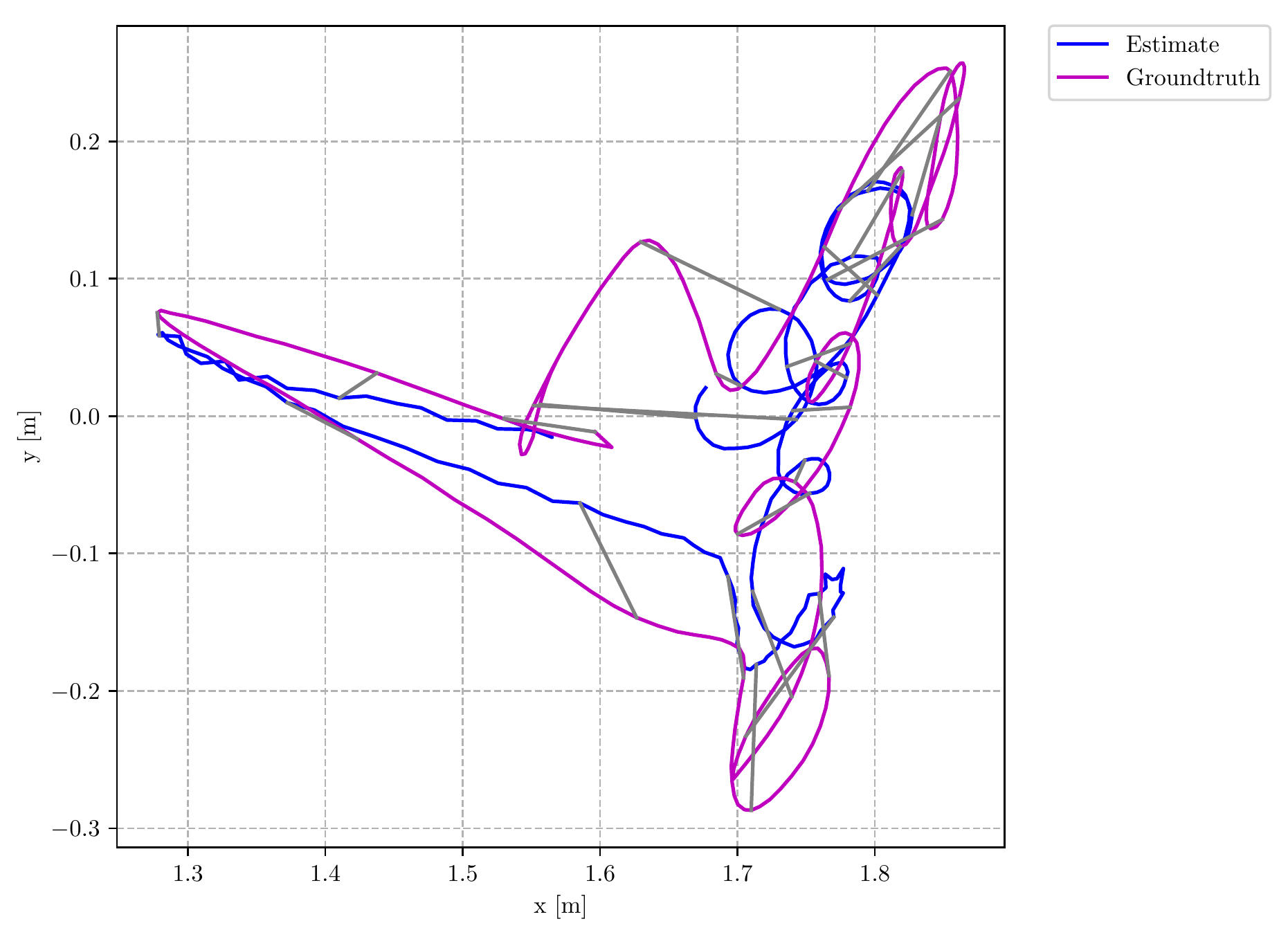}\hfill
    \includegraphics[width=0.33\linewidth, height=4cm, trim=0.0 0.0 0.0 0.0, clip]{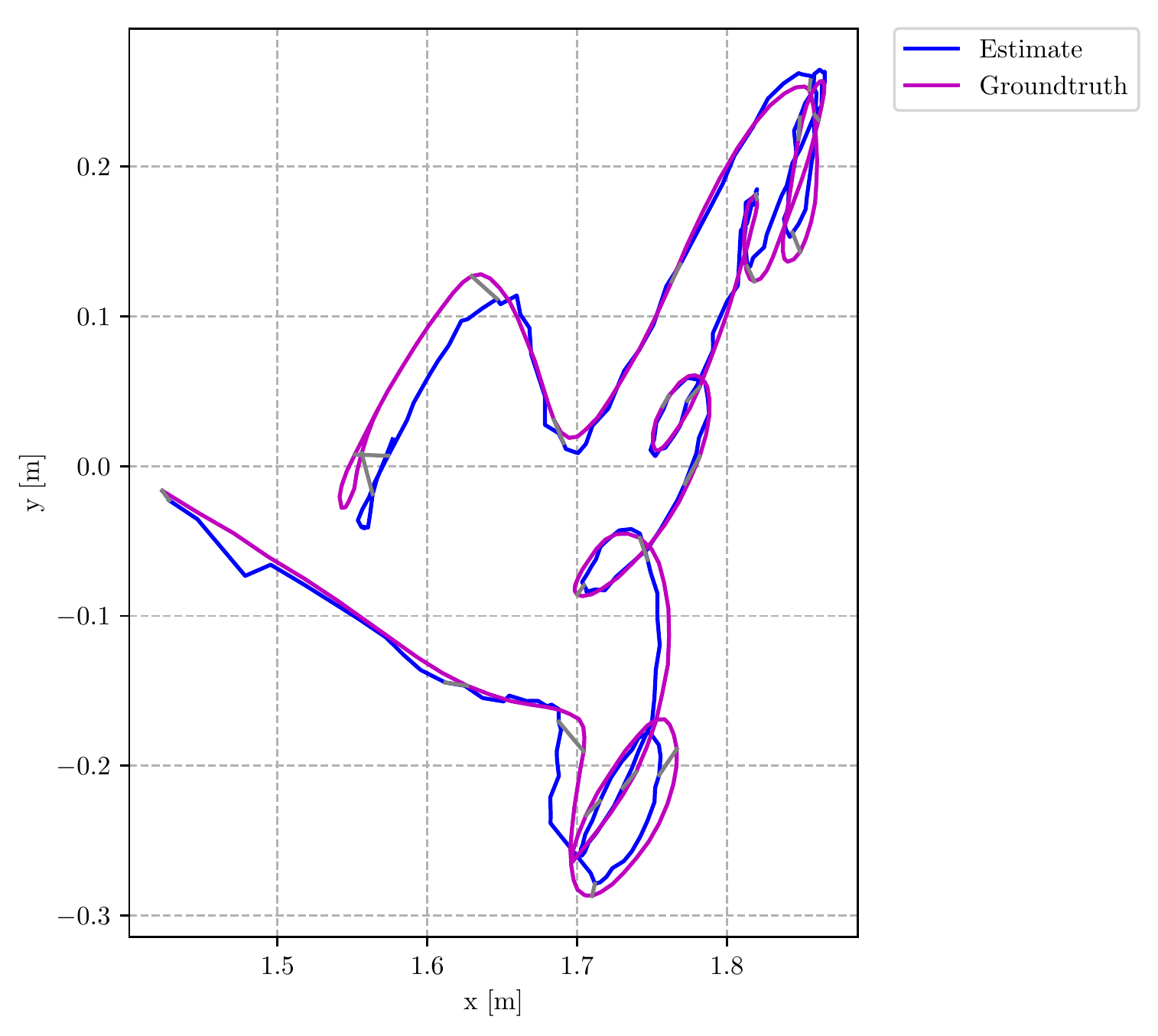}\hfill
    \includegraphics[width=0.33\linewidth, height=4cm, trim=0.0 0.0 0.0 0.0, clip]{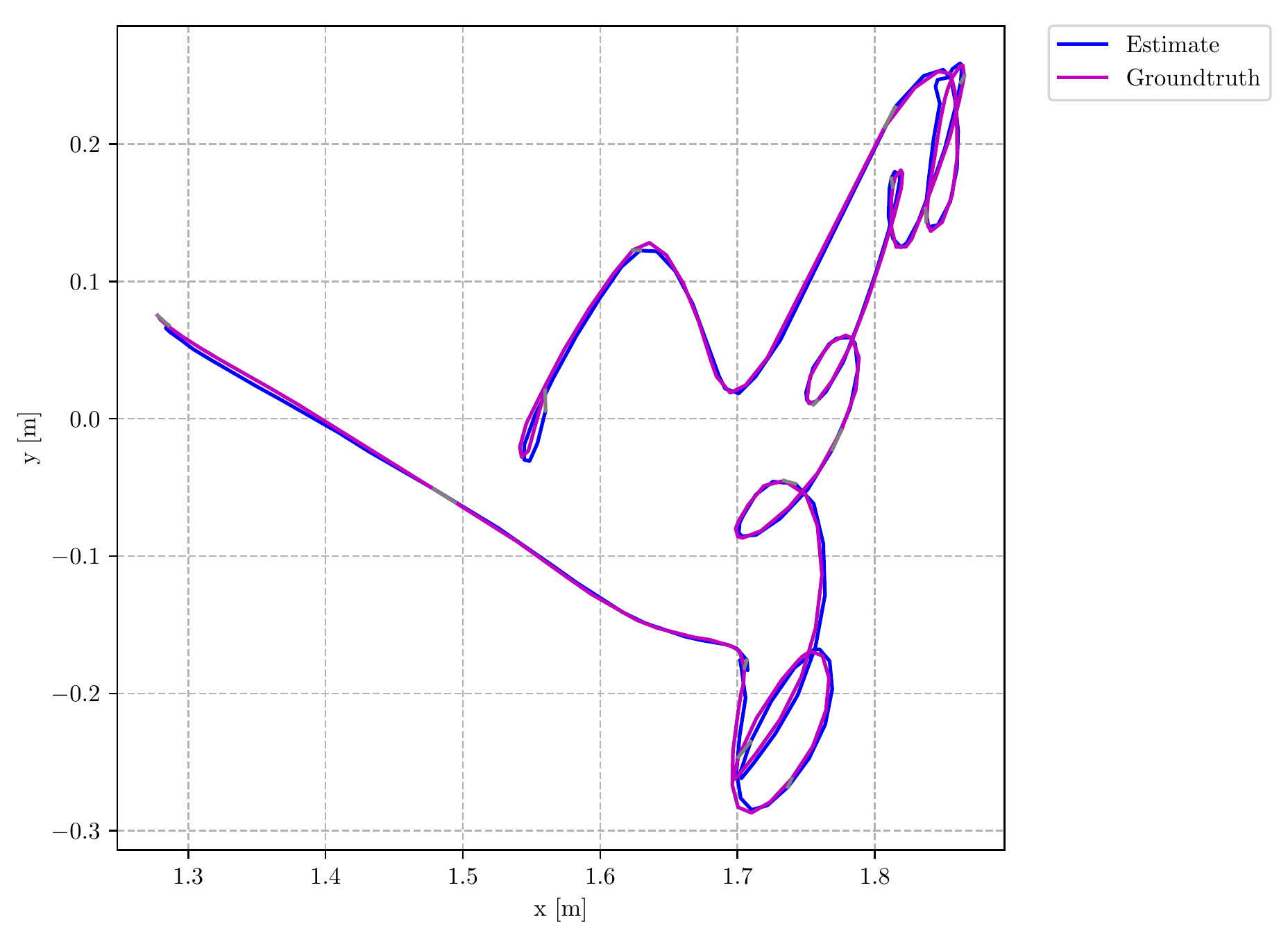}\hfill
\vspace{-0.1cm}
\captionof{figure}{\label{fig:desk:trajectories}
        DSO (left), ORB-SLAM (center) and EDS (right) camera trajectory
        for sequence~\rpgdesk{} in the dataset from~\cite{Zhou18eccv} at~20~FPS.
}
\vspace{2.5ex}
\begin{adjustbox}{max width=\linewidth}
\setlength{\tabcolsep}{2pt}
\begin{tabular}{lcccccccccccccccc}
        & \multicolumn{4}{c}{20 FPS} & \multicolumn{4}{c}{10 FPS} &
        \multicolumn{4}{c}{7 FPS} & \multicolumn{4}{c}{5 FPS} \\

        \cmidrule(l{1mm}r{1mm}){2-5} \cmidrule(l{1mm}r{1mm}){6-9}
        \cmidrule(l{1mm}r{1mm}){10-13} \cmidrule(l{1mm}r{1mm}){14-17}

    & DSO* & DSO & ORB\_SLAM* & EDS (\textbf{ours}) &
    DSO* & DSO & ORB\_SLAM* & EDS(\textbf{ours}) &
    DSO* & DSO & ORB\_SLAM* & EDS(\textbf{ours}) &
    DSO* & DSO & ORB\_SLAM* & EDS(\textbf{ours}) \\

Input   & F     & F     & F    & E+F & F     & F    & F     & E+F & F     & F    & F    & E+F & F     & F    & F    & E+F \\
\midrule

    \multicolumn{1}{l|}{\rpgbin} &
    1.1   & 1.2   & 2.4  & \textbf{1.1} & %
    1.8   & 1.8  & 2.4   & \textbf{1.8} & %
    3.5   & 2.5  & \textbf{2.4}  & 2.5 & %
    16.9  & 4.8  & \textbf{2.5}  & 2.6 \\ %
    \multicolumn{1}{l|}{\rpgboxes} &
    \textbf{2.0}   & 2.0   & 3.9  & 2.1 &
    13.5  & \textbf{3.1}  & 3.9   & 3.8 &
    14.8  & 14.6 & \textbf{3.9}  & 5.0 &
    14.8  & 14.6 & 7.0  & \textbf{5.8} \\
\multicolumn{1}{l|}{\rpgdesk} &
    10.0  & 10.0  & 3.8  & \textbf{1.5} &
    13.4  & 9.1  & 3.8   & \textbf{3.4} &
    21.1  & 16.2 & 7.8  & \textbf{4.7} &
    21.6  & 19.2 & 9.3  & \textbf{5.0} \\
\multicolumn{1}{l|}{\rpgmonitor} &
    \textbf{0.9}   & 0.9   & 3.1  & 1.0 &
    3.9   & \textbf{1.5}  & 10.6  & 2.3 &
    26.5  & 12.1 & 10.9 & \textbf{2.5} &
    28.0  & 27.1 & 10.3 & \textbf{8.0}\\
\bottomrule
\end{tabular}
\end{adjustbox}
    \captionof{table}{\label{tab:stereo_davis_study}Performance at different frame rates in terms of Absolute Trajectory Error (RMS)~[t:~\si{\centi\meter}].
    Data from \cite{Zhou18eccv}. }
\end{figure*}

\hyphenation{separately}
\hyphenation{came-ra}
\hyphenation{USLAM}

\section*{Overview}
\label{sec:suppl:overview}

In this supplementary material we present:
\begin{itemize}%

    \item Additional details on the low frame rate experiments (\cref{sec:experimdetails}).

    \item Qualitative results about the sensitivity study in depth inaccuracies (\cref{sec:sensitivity:depth}).

    \item Our beamsplitter device and two additional experiments recorded with it (\cref{sec:supexperim}).

    \item A novel dataset with high quality events \& RGB frames recorded with 
        the beamsplitter device to foster research on the topic
        (\cref{sec:eds:dataset}).

    \item A more thorough discussion of limitations (\cref{sec:suppl:limitations}).

\end{itemize}

\section{Low Frame Rate Experiments}
\label{sec:experimdetails}

\Cref{tab:stereo_davis_study} shows the numerical values in the low frame rate
experiment, used for plotting \cref{fig:fps_study}.
The values shows that
EDS performs the best when the number of frames decreases.  Direct methods are
less accurate than indirect methods at a lower frame rate (FPS).  However at a
higher frame rate direct methods use more information from the scene; achieving
a higher accuracy.

Among all sequences in dataset~\cite{Zhou18eccv}, \rpgdesk{} has the most
challenging camera motions.  In this scenario events excel and help standard
frame-based methods.  This is also shown in the results (see
\cref{tab:stereo_davis_study}) where EDS outperforms previous frame-based
methods at any given frame rate.  \Cref{fig:desk:trajectories} shows the
trajectories of DSO, ORB-SLAM and EDS for the~\rpgdesk{} sequence.  It is
qualitatively visible how events enhance standard direct methods guiding the
camera pose and producing a more accurate trajectory.

\section{Sensitivity with respect to Depth Noise}
\label{sec:sensitivity:depth}
\begin{figure*}[t]
	\centering
    {\small
    \setlength{\tabcolsep}{2pt}
	\begin{tabular}{
	>{\centering\arraybackslash}m{0.4cm}
	>{\centering\arraybackslash}m{0.25\linewidth}
	>{\centering\arraybackslash}m{0.25\linewidth}
	>{\centering\arraybackslash}m{0.4\linewidth}}
		& 
		Front view & 
		Top view & 
		Trajectory, top view\\[1ex]%

        \rotatebox{90}{\makecell{1\%}}&
		\wframe{\includegraphics[width=\linewidth]{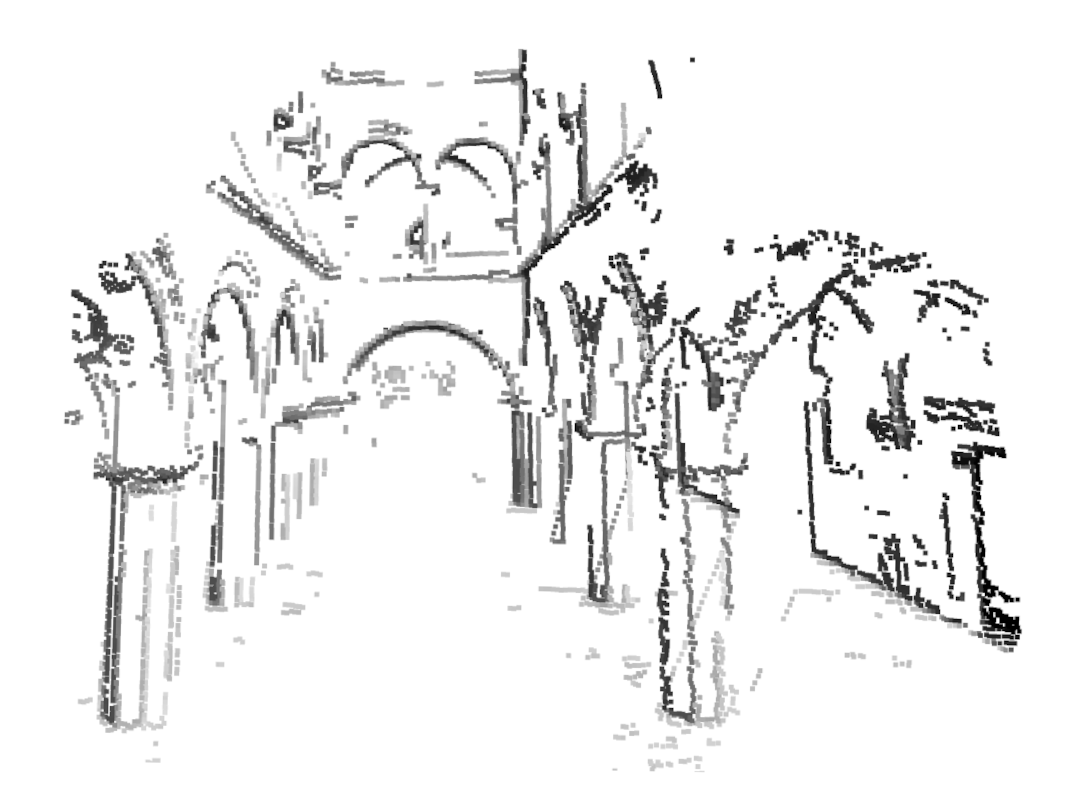}}&
		\wframe{\includegraphics[angle=180,origin=c,width=\linewidth]{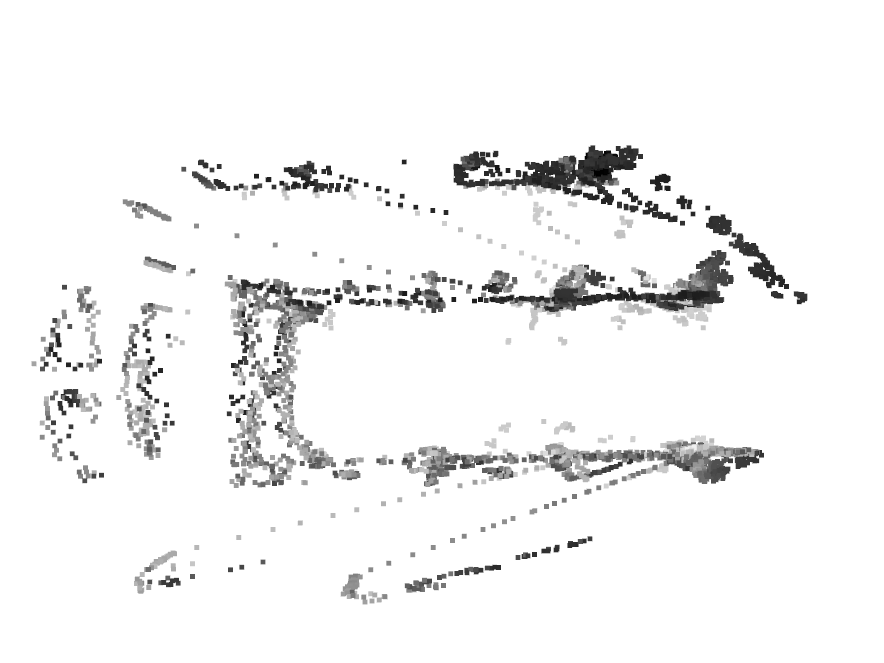}}&
		\wframe{\includegraphics[trim={0 0 3.65cm 0},clip,angle=270,width=\linewidth]{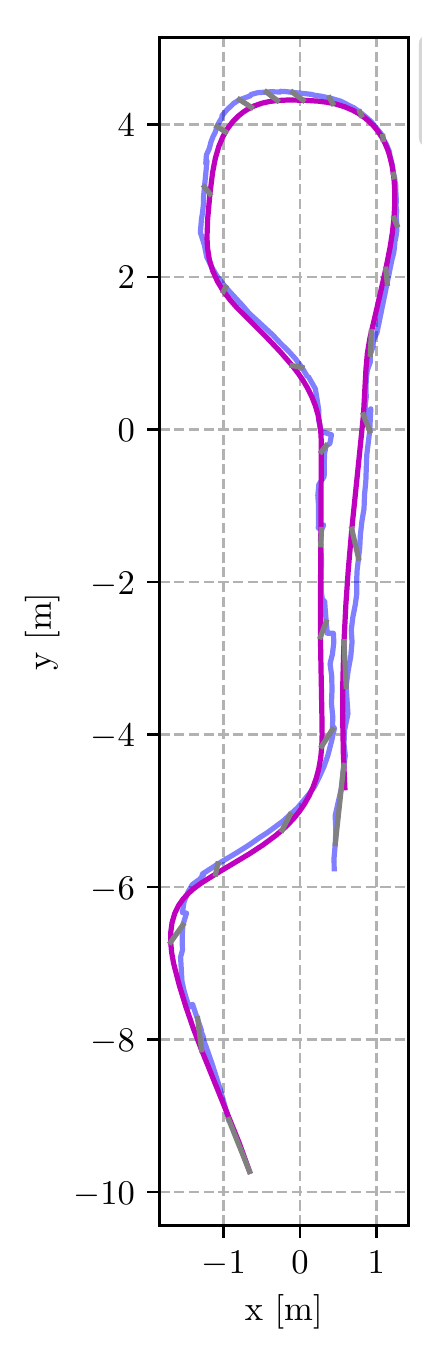}}
		\\

        \rotatebox{90}{\makecell{5\%}}&
		\wframe{\includegraphics[width=\linewidth]{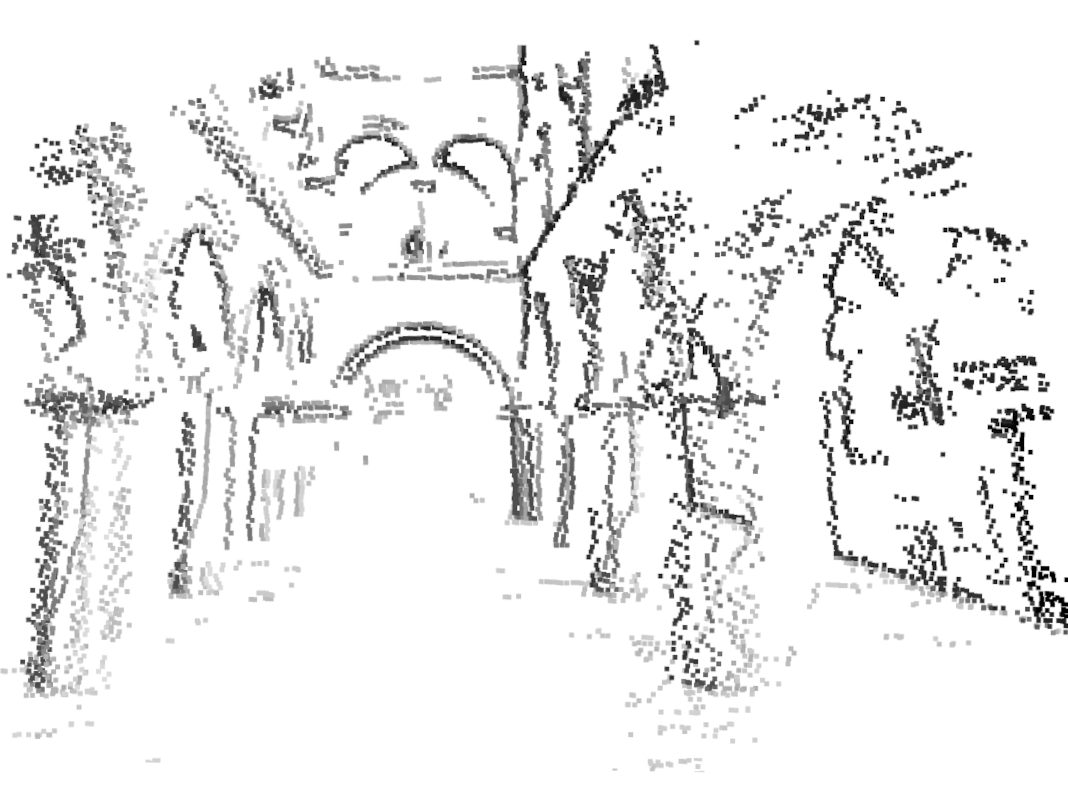}}&
		\wframe{\includegraphics[angle=180,origin=c,width=\linewidth]{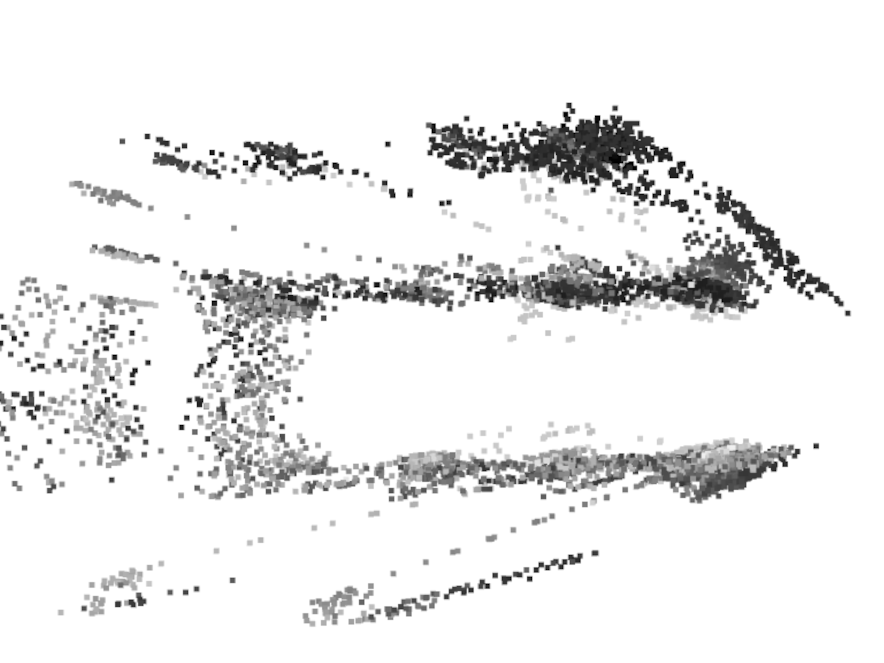}}&
		\wframe{\includegraphics[trim={0 0 3.65cm 0},clip,angle=270,width=\linewidth]{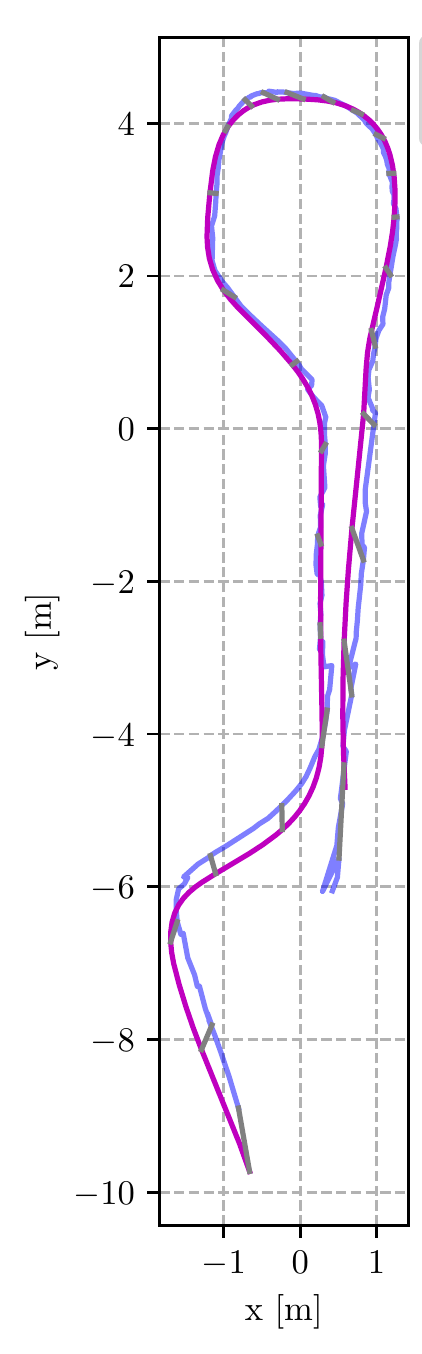}}
		\\

        \rotatebox{90}{\makecell{10\%}}&
		\wframe{\includegraphics[width=\linewidth]{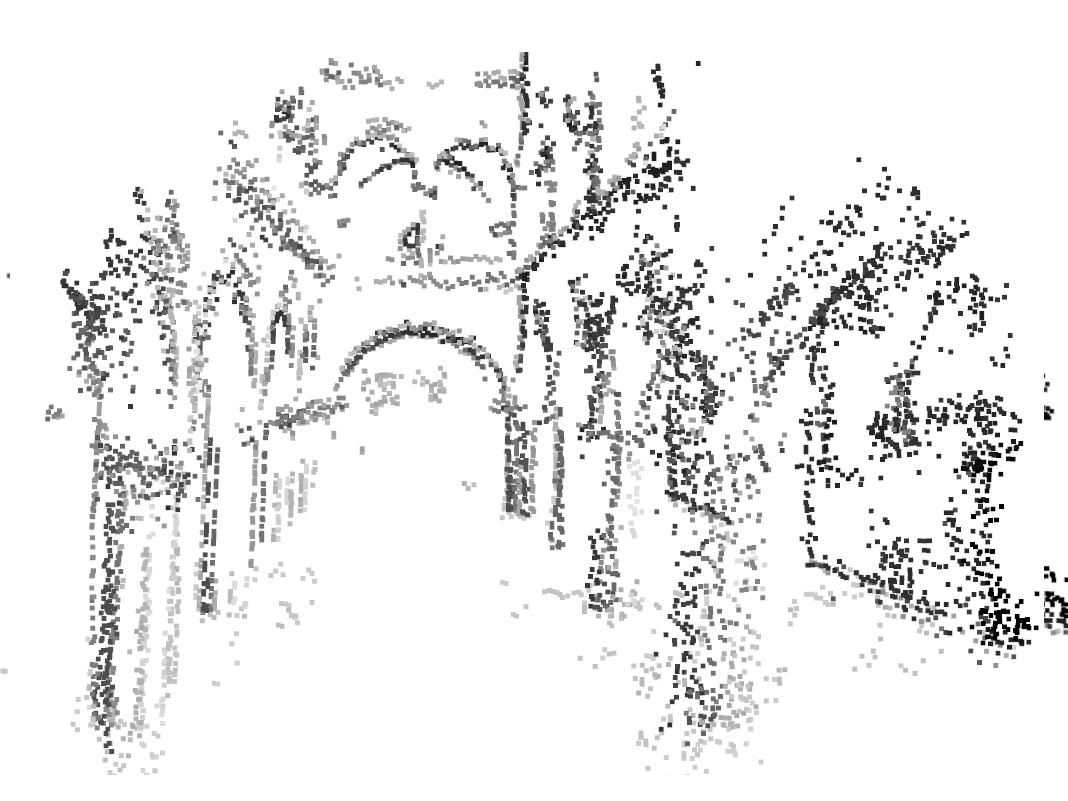}}&
		\wframe{\includegraphics[angle=180,origin=c,width=\linewidth]{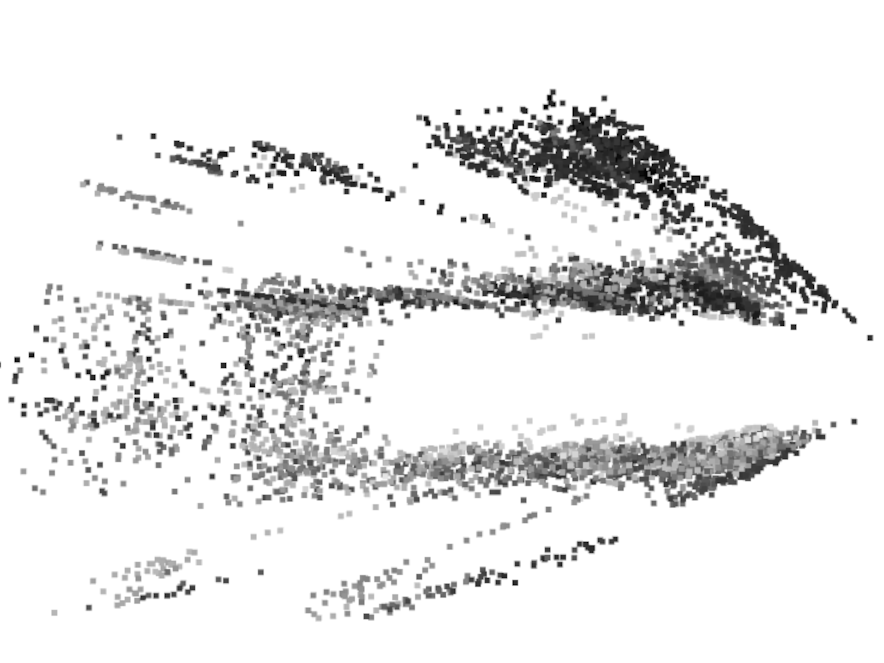}}&
		\wframe{\includegraphics[trim={0 0 3.65cm 0},clip,angle=270,width=\linewidth]{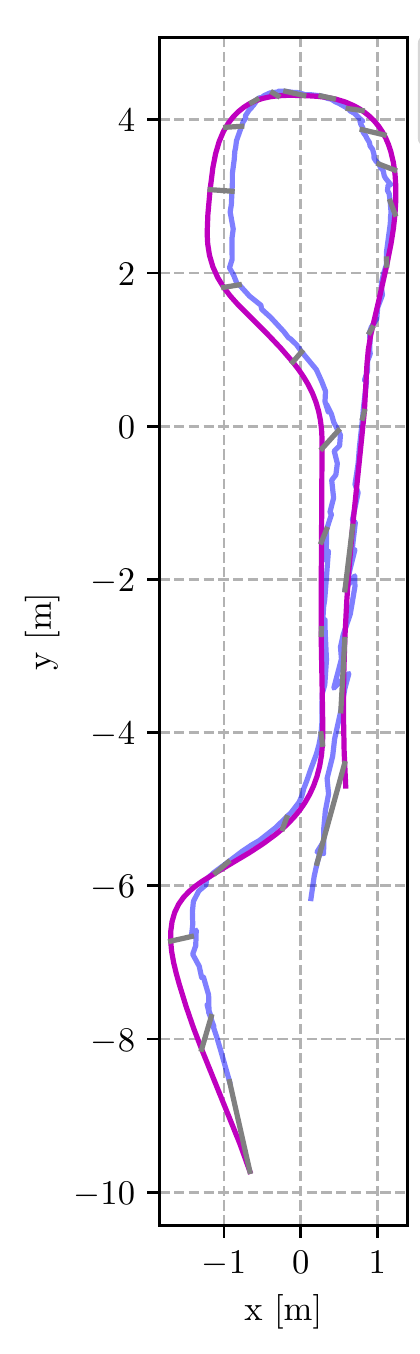}}
		\\

        \rotatebox{90}{\makecell{20\%}}&
		\wframe{\includegraphics[width=\linewidth]{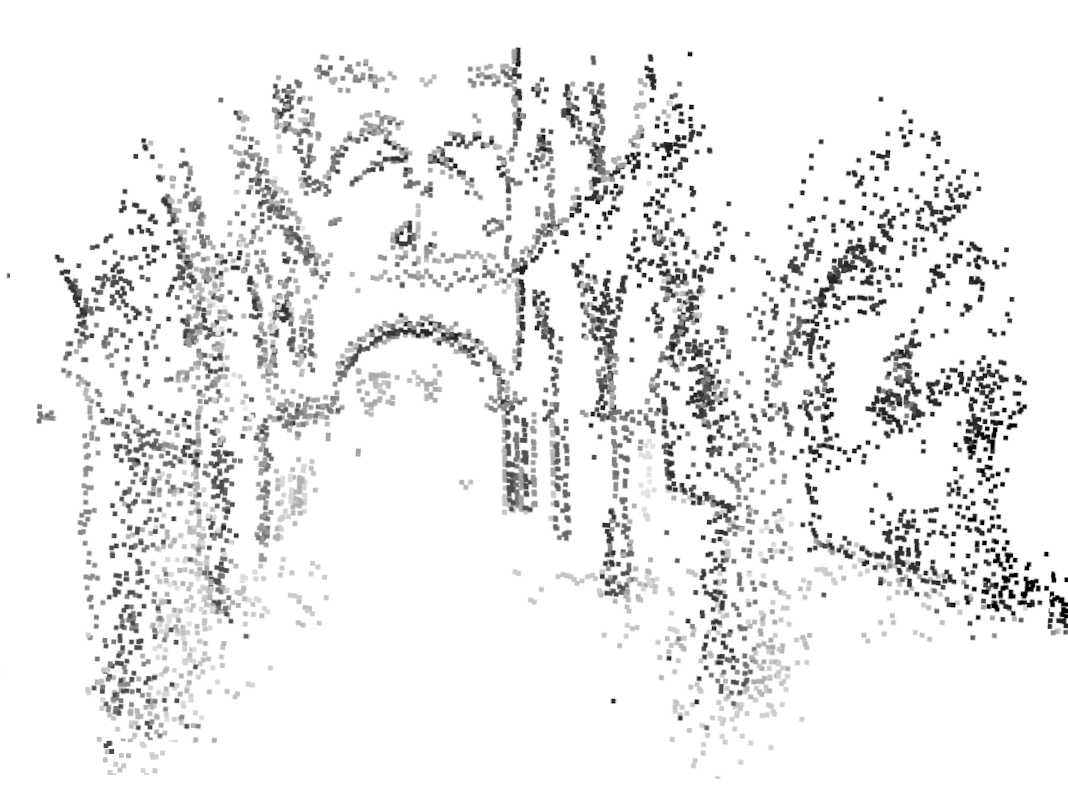}}&
		\wframe{\includegraphics[angle=180,origin=c,width=\linewidth]{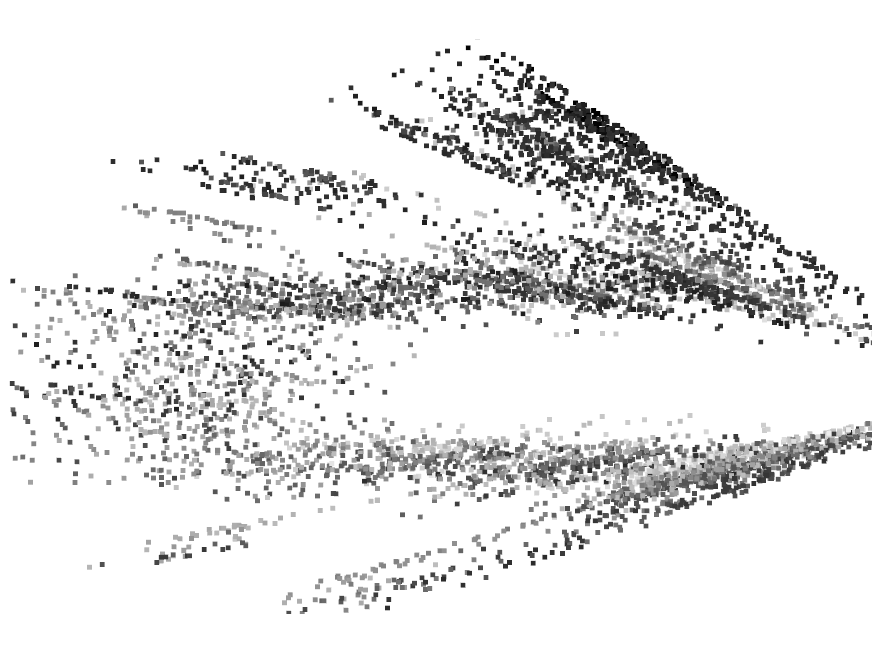}}&
		\wframe{\includegraphics[trim={0 0 3.65cm 0},clip,angle=270,width=\linewidth]{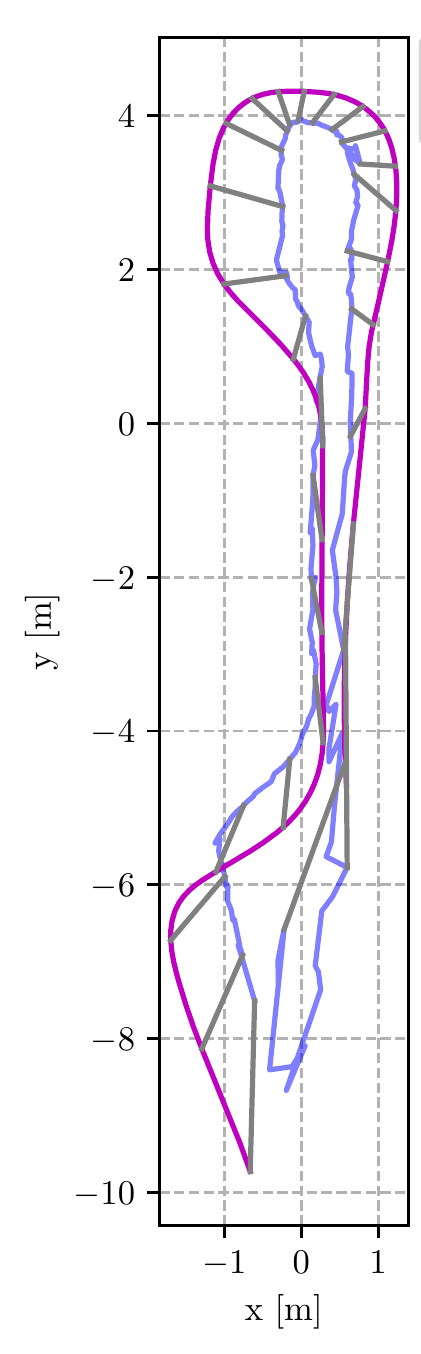}}
		\\
		
	\end{tabular}
	}
	\vspace{-1ex}
    \caption{Sensitivity with respect to increasing noise in the depth of the 3D points, from 1\% to 20\% of the median scene depth. 
    Atrium scene.
    The ground truth trajectory is in purple, while the estimated trajectory is in blue.
    }
	\label{fig:atriumpointclouds}
	\vspace{-1ex}
\end{figure*}

Depth estimation is the main limitation of our EDS method. This is because events are only
used for camera tracking (front-end) and the EGM highly depends on optical
flow, which is a function of depth. We previously presented how depth noise affects the
camera tracker and entails camera pose inaccuracies. 
\Cref{fig:atriumpointclouds} shows the Atrium sequences with the perturbed map. 
We might conclude that up to \si{10~\percent} standard deviation
of the median depth in the scene, events are still useful to track the camera motion.
However, when the mapper wrongfully estimates depth points, events do not help the
tracker, generating more inaccuracies than if those events were not used.
Therefore, it is important to correctly select which points to incorporate in the EGM calculation.

\section{Beamsplitter and Additional Experiments}
\label{sec:supexperim}

Publicly available event-based Visual Odometry (VO) datasets are limited and/or
contain low quality sensory data~\cite{Mueggler17ijrr,Zhou18eccv}, either
because of noisy events recorded with pioneer neuromorphic devices (such as the
DAVIS240B~\cite{Brandli14ssc} or DAVIS346B) or due to low quality frames (i.e.,
no gamma correction, small fill factor).  Recent publicly available datasets,
such as~\cite{Gehrig21raldsec,Klenk21iros}, are equipped with newer sensors,
however they do not share the same optical axis, having frames and events in two
image planes, standard and event-based camera separately. There is a clear need
for up-to-date datasets in the event-based VO community with state-of-the-art
sensors.  We now evaluate our monocular VO method in an extended dataset with
our custom co-capture device. First, we present our custom-made beamsplitter device
and the calibration procedure in brief~(\cref{sec:supexperim:tracking}).
Second, we evaluate the performance of the method (\cref{sec:supexperim:bs}).

\subsection{Beamsplitter: Sensors and Calibration}
\label{sec:supexperim:tracking}

\begin{figure}[t]
    \centering
    \includegraphics[width=\linewidth,trim={1.0cm 0.0cm 0 0},clip]{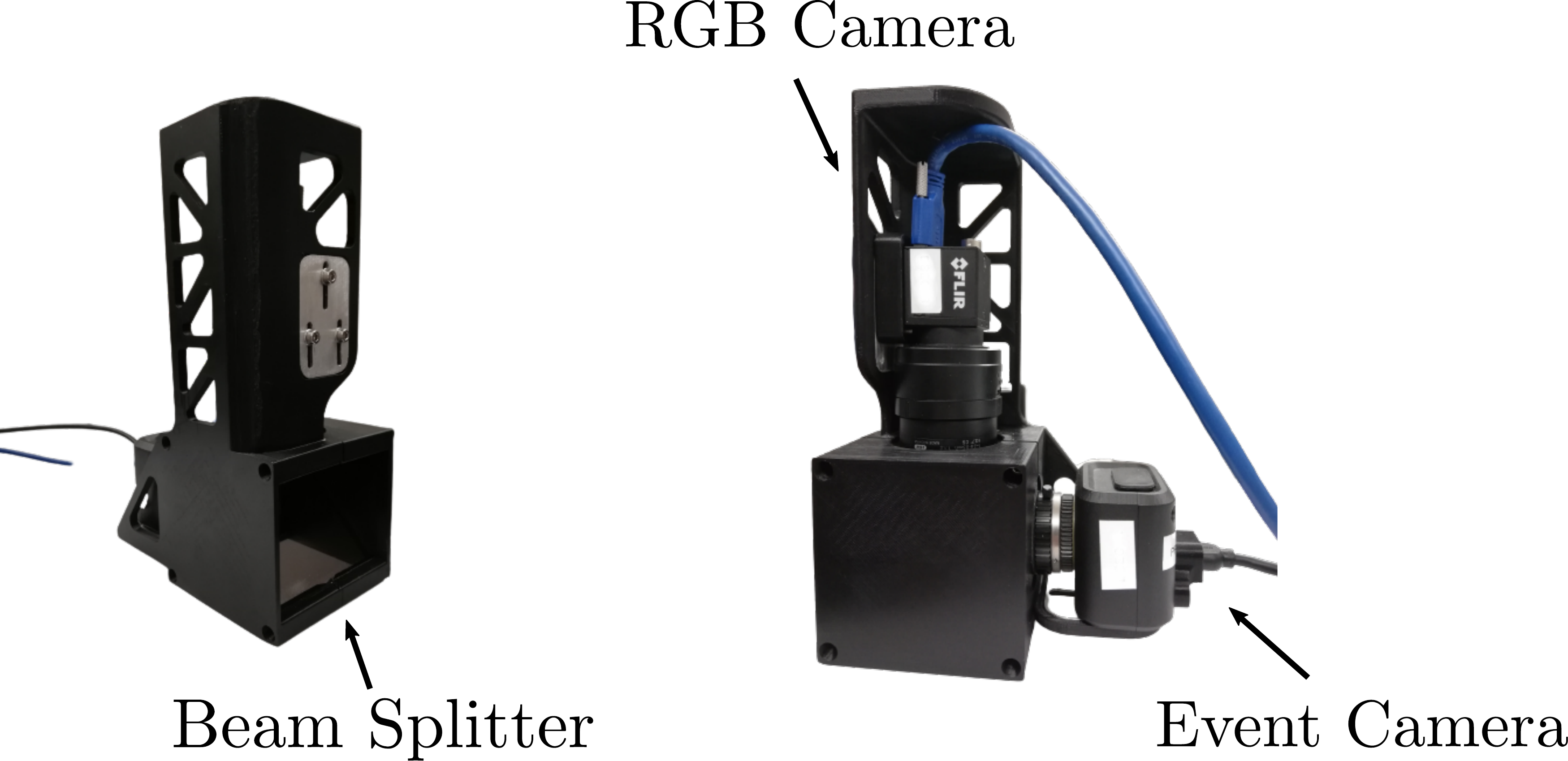}
    \caption{Beamsplitter with a Prophesee Gen3 event camera and a FLIR color camera. Mounted
    on a case with a 50R/50T beam splitter mirror that allows the sensors
    to share a spatially aligned field of view.
    }
    \label{fig:beamsplitter}
\end{figure}

We build a custom-made sensing device consisting of a Prophesee Gen3 event
camera~\cite{propheseeevk} ($640 \times 480$ px resolution) and a color FLIR
camera viewing the same scene through a
beamsplitter. The Snoopy house scene in \cref{fig:eyecatcher} is recorded
with such a device (see \cref{fig:beamsplitter} and \cref{tab:dataset:sensor}
for details).

\begin{figure}[b]
  \centering
  \frame{\includegraphics[width=0.49\linewidth]{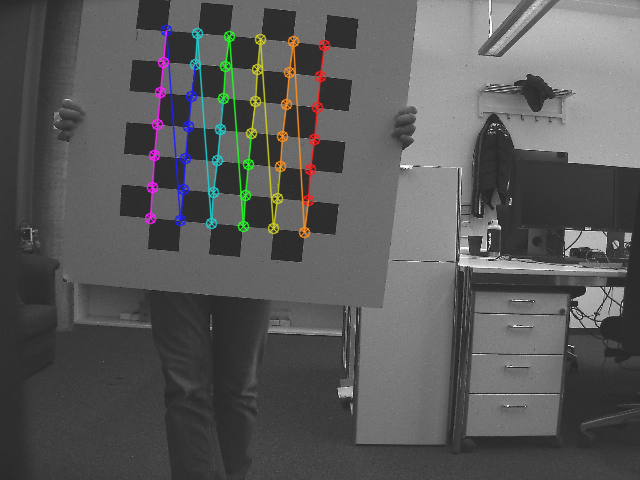}}\hfill
  \frame{\includegraphics[width=0.49\linewidth]{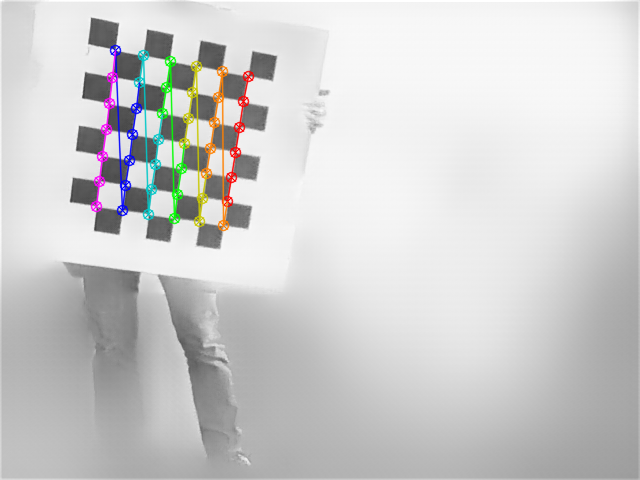}}
  \caption{Calibration. Frames from the standard FLIR camera (top) and 
  grayscale frames reconstructed from events (bottom), 
  both with checkerboard corners used during calibration.}
  \label{fig:beamsplitter:calib}
\vspace{-2ex}
\end{figure}

\begin{figure}[t]
    \centering
    \frame{\includegraphics[width=0.99\linewidth, trim={1.0cm 3.0cm 6.0cm 3.0cm},clip]{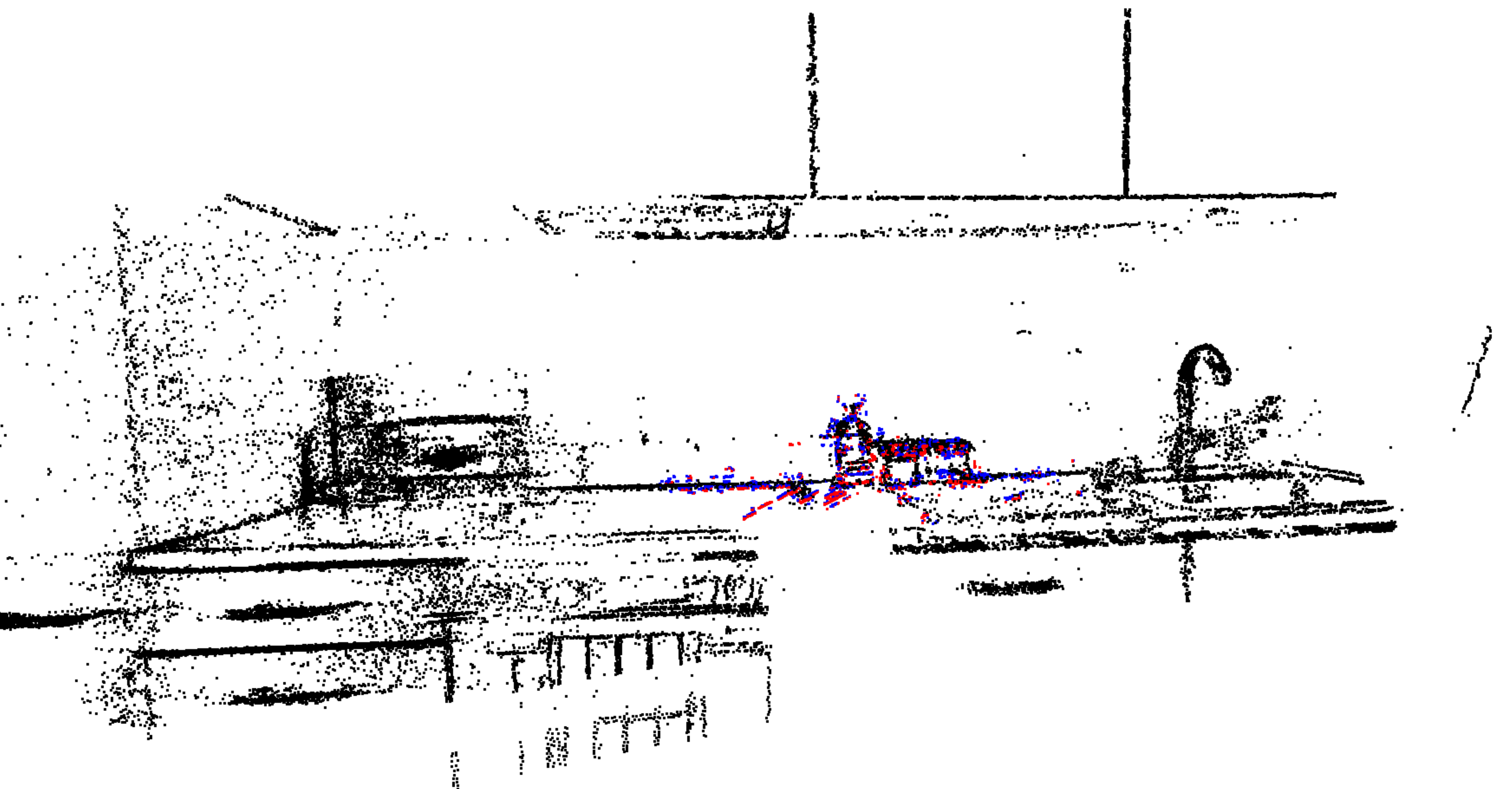}}\\[0.5ex]
    \frame{\includegraphics[width=0.99\linewidth, trim={3.0cm 0.0cm 1.0cm 0.0cm},clip]{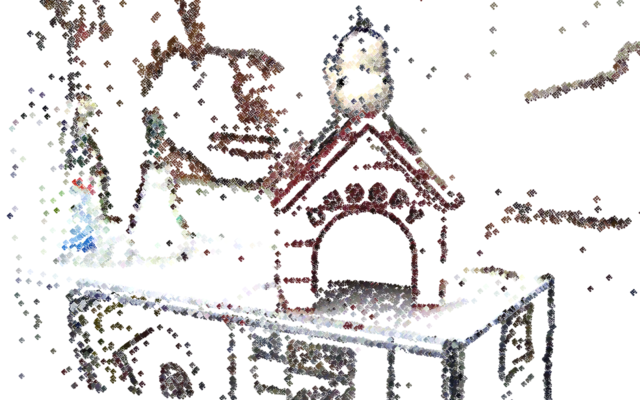}}
    \caption{
        Point cloud maps of the kitchen (top) and snoopy (bottom) sequences.
    }
    \label{fig:sequences:trajectories}
\end{figure}

\begin{figure}[tb]
    \centering
    \ifcompileanimated
     \frame{\animategraphics[width=0.99\linewidth]{10}{imgs/kitchen/frame_00000004}{20}{90}}
     \else
     \frame{\includegraphics[width=0.99\linewidth]{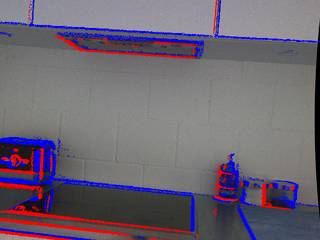}}
     \fi
    \caption{Beamsplitter output with aligned events and frames for the
    \rpgkitchen{} sequence. This figure contains animations that can be viewed in \emph{Acrobat Reader}.}
    \label{fig:kitchen_anim}
    \vspace{-2ex}
\end{figure}

Both cameras are calibrated and their outputs are aligned with sub-pixel
accuracy, giving the equivalent of a DAVIS camera with higher quality color
frames at VGA ($640 \times 480$ px) resolution. To calibrate the cameras, we
reconstruct grayscale frames from events using e2calib~\cite{Muglikar21cvprw}
and input them into Kalibr~\cite{Rehder16icra}, which computes the intrinsic
and extrinsic camera parameters. We calibrate each camera separately to estimate
the lens distortion parameters and focal lengths. We configure the FLIR camera to
acquire frames at VGA resolution. 
The color image is warped into the event camera using extrinsic calibration with sub-millimeter accuracy. 
The intrinsic calibrations of both
cameras have reprojection errors of approximately \SI{0.2}{pixels}. 
This gives a sufficiently good alignment
into one single field of view image.
\Cref{fig:beamsplitter:calib} depicts
individual camera outputs during the calibration process,
and~\cref{fig:kitchen_anim} shows the rectified and undistorted image output
with events aligned.

\subsection{Ego-motion Estimation Results}
\label{sec:supexperim:bs}

We recorded two sequences with our beamsplitter device to prove
generalization of our method to other (and newer) visual sensory data.
The sequences were recorded in natural indoor scenes and we used
COLMAP~\cite{Schoenberger16cvpr}
to provide a ground truth trajectory, since external motion-capture system
(i.e., Vicon or Optitrack) was not available.~\Cref{fig:sequences:trajectories}
shows the reconstructed 3D map for EDS in the \rpgkitchen{} and \rpgsnoopy{}
sequences using events and frames. The colored points in the central part of the
kitchen correspond to the 3D locations which are active (i.e., generating events
using the event generation model - EGM) in the current keyframe.~\rpgsnoopy{}
points cloud used RGB color for a more appealing visualization of the Snoopy
reconstructed house.

\begin{table}[t]
\centering
\begin{adjustbox}{max width=\linewidth}
\setlength{\tabcolsep}{6pt}
\begin{tabular}{llcccc}
    & & ORB-SLAM \cite{MurArtal17tro} & EVO \cite{Rebecq17ral} & DSO \cite{Engel17pami}  & EDS (\textbf{Ours})\\
    & \textbf{Input} & F & E & F & E+F\\ 
\midrule
\multirow{4}{*}{\rotatebox[origin=c]{90}{\textbf{}}}
    & \rpgkitchen{}     & 13.0$\pm$9.7 & - & 12.5$\pm$6.8 & \textbf{9.6$\pm$5.5} \\
    & \rpgsnoopy{}     & 35.1$\pm$28 & - & \textbf{30.7$\pm$14} & 30.9$\pm$13 \\
\bottomrule
\end{tabular}
\end{adjustbox}
\caption{\label{tab:compare:appendix}
    Comparison with state-of-the-art 6-DOF VO methods in terms of Absolute
    Trajectory Error (RMS)~[t:~\si{\centi\meter}] and its standard deviation 
    on two beamsplitter sequences. 
    Ground truth poses are computed using COLMAP.
    Input data may be: events (E) and/or grayscale frames (F). 
    EVO entries marked with hyphen indicate the code did not manage to recover poses on these sequences.
}
\end{table}

\def\figWidth{0.24\linewidth}
\begin{figure*}[!ht]
\centering
{\small
\setlength{\tabcolsep}{2pt}
\begin{tabular}{
>{\centering\arraybackslash}m{\figWidth} 
>{\centering\arraybackslash}m{\figWidth}
>{\centering\arraybackslash}m{\figWidth}
>{\centering\arraybackslash}m{\figWidth}}

    \frame{\includegraphics[width=\linewidth]{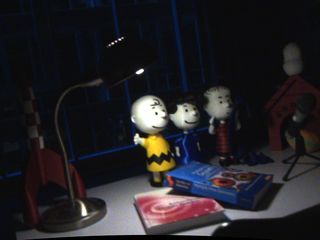}}&
    \frame{\includegraphics[width=\linewidth]{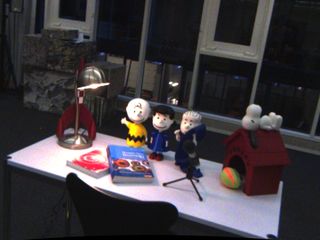}}&
    \frame{\includegraphics[width=\linewidth]{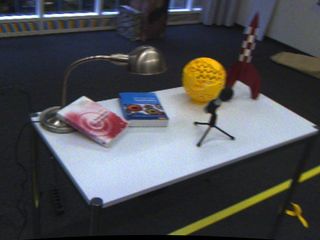}}&
    \frame{\includegraphics[width=\linewidth]{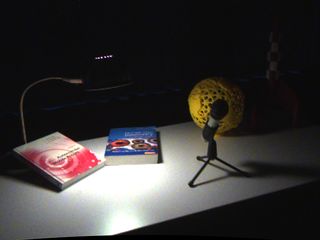}}
    \\

    \frame{\includegraphics[width=\linewidth]{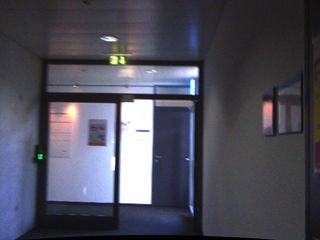}}&
    \frame{\includegraphics[width=\linewidth]{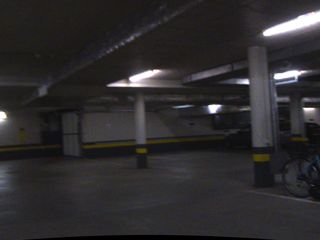}}&
    \frame{\includegraphics[width=\linewidth]{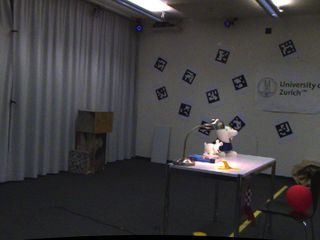}}&
    \frame{\includegraphics[width=\linewidth]{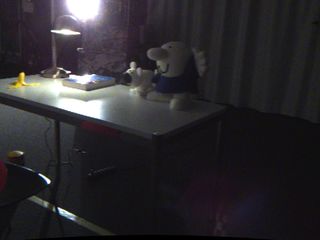}}
    \\

    \frame{\includegraphics[width=\linewidth]{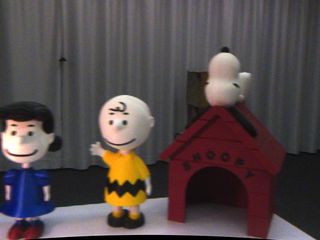}}&
    \frame{\includegraphics[width=\linewidth]{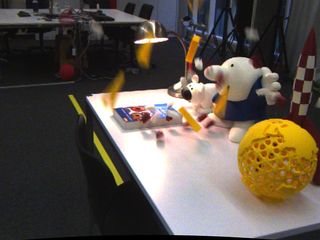}}&
    \frame{\includegraphics[width=\linewidth]{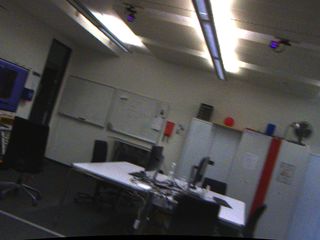}}&
    \frame{\includegraphics[width=\linewidth]{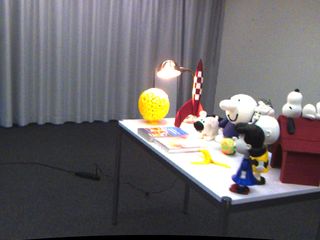}}
    \\

    \frame{\includegraphics[width=\linewidth]{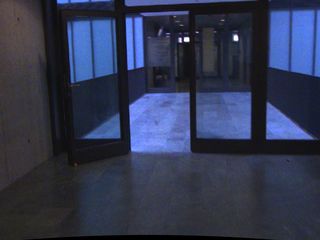}}&
    \frame{\includegraphics[width=\linewidth]{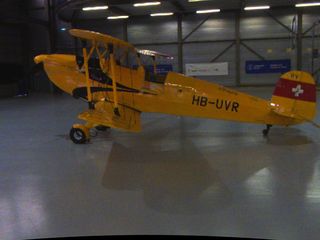}}&
    \frame{\includegraphics[width=\linewidth]{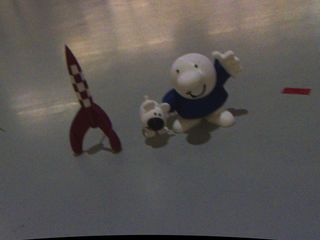}}&
    \frame{\includegraphics[width=\linewidth]{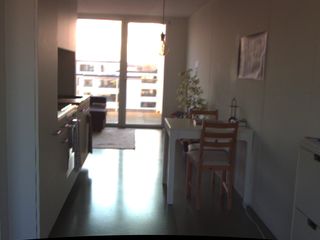}}
    \\
\end{tabular}
}
\caption{\emph{Sequences from our EDS Dataset}: 
    the dataset contains 16 sequences with events, color frames, IMU data and ground truth poses 
    on a diverse set of environments.}
	\label{fig:eds:dataset}
\end{figure*}

Quantitatively (\cref{tab:compare:appendix}), our method outperforms all other
monocular pure VO baseline methods in the~\rpgkitchen{} sequence and achieves
state-of-the-art accuracy in the~\rpgsnoopy{} sequence.  We did not include
comparison with USLAM~\cite{Rosinol18ral} in~\cref{tab:compare:appendix} since
it requires an IMU, which is not available in these two sequences.  The
beamsplitter device (see \cref{sec:supexperim:bs}) has a better sensor quality
than the DAVIS240C in~\cite{Zhou18eccv}. It produces events with less noise,
which makes the EGM more accurate (see multimedia material for details).

\section{The EDS Dataset}
\label{sec:eds:dataset}
We realized the existing gap in good quality monocular visual-inertial odometry
(VIO) dataset with events.  Therefore we built a beamsplitter device during the
last part of our investigation.  The aim of the beamsplitter was to record and
release a new dataset to promote and facilitate research on the topic.  The
dataset targets VIO but it may be used to demonstrate other tasks, such as
optical flow estimation, depth estimation, view synthesis (e.g., NeRFs), etc.
\Cref{fig:eds:dataset} shows snapshots of the recorded sequences with our
beamsplitter device.  When available, we also record ground truth poses from a
motion-capture system.

In a nutshell, the \href{https://rpg.ifi.uzh.ch/eds.html#dataset}{EDS dataset}
provides synchronized events with RGB frames, IMU
and ground truth data. The data is given in three different formats,
pocolog\footnote{\url{https://www.rock-robotics.org}},
rosbag\footnote{\url{http://wiki.ros.org/rosbag}} and archive files with
compressed events in HDF5 format. 
The images are timestamped, with exposure time and gain values, 
and the ground truth poses are at the camera frame 
(i.e., $T_{\text{marker}\_\text{cam}}$ already applied). 
The motion capture system is Vicon or Optitrack depending on the scene. 
Long sequences such as \href{https://rpg.ifi.uzh.ch/eds.html#04_floor_loop}{\edsfloorloop{}},
\href{https://rpg.ifi.uzh.ch/eds.html#05_rpg_building}{\edsrpgbuilding{}},
and \href{https://rpg.ifi.uzh.ch/eds.html#12_floor_eight_loop}{\edsflooreightloop{}}
provide ground truth at the start and end locations. 
\href{https://rpg.ifi.uzh.ch/eds.html#15_apartment_day}{\edsaptday{}} gives start and finish positions using an
Apriltag\footnote{\url{https://april.eecs.umich.edu/software/apriltag}} marker.
The \href{https://rpg.ifi.uzh.ch/eds.html#calibration}{calibration} results to align frames
and events as well as the
camera to IMU transformation are given.

\begin{table}[t]
\centering
\begin{adjustbox}{max width=\linewidth}
\setlength{\tabcolsep}{6pt}
\begin{tabular}{l|l}
\textbf{Sensor Type} & \textbf{Description}\\
\midrule
Prophesee Gen 3.1 &
    \begin{tabular}[c]{@{}l@{}}Prophesee PPS3MVCD event camera\\
    640 $\times$ 480 pixels, 3/4” CMOS Monochrome\\
    HFOV 59.4$^\circ$ VFOV 34.7$^\circ$ \\
    $\geq$120 dB dynamic range\end{tabular} \\
\midrule
FLIR Blackfly S USB3 &
    \begin{tabular}[c]{@{}l@{}}FLIR BFS-U3-16S2C-CS\\
    640 $\times$ 480 pixels, 1/2.9” CMOS Color\\
    HFOV 45.31$^\circ$ VFOV 34.7$^\circ$ \\
    71.4 dB dynamic range\\
    up to 75 FPS (depending on the sequence)
    \end{tabular} \\
\midrule
Inversense MPU-9250 &
    \begin{tabular}[c]{@{}l@{}}Inertial Measurement Unit\\ MEMS, 16bits resolution\\
    3~$\times$ Gyroscopes\\
    3~$\times$ Accelerometers\\
    3~$\times$ Magnetometers (not utilized)\\
    1000Hz sampling rate
    \end{tabular} \\
\midrule
Motion-capture system & \begin{tabular}[c]{@{}l@{}}Optitrack: RPG Flying room, 11 cameras\\
Vicon: RPG Drone Arena, 36 cameras\\
Camera pose: position + rotation \\
150 Hz sampling rate\end{tabular}\\
\bottomrule
\end{tabular}
\end{adjustbox}
\caption{
    Details of the hardware utilized for the dataset collection
}
\label{tab:dataset:sensor}
\vspace{-2ex}
\end{table}

\section{Limitations}
\label{sec:suppl:limitations}

\subsection{Grayscale Frames are not HDR}
Our method works under the assumption of the availability of collocated
grayscale frames and events. Current devices, such as the DAVIS
camera~\cite{Brandli14ssc}, produce low-quality grayscale frames, with a dynamic
range of $\approx$~\SI{55}{\decibel}, which is small compared to the high
dynamic range (HDR) properties of event cameras ($>$\SI{120}{\decibel}). 
Hence, the frames from the DAVIS are not HDR, and one could point this as a limitation
of the approach.

We tackled this problem in two ways: ($i$) by building our own sensing device
with higher quality frames than the DAVIS (\cref{fig:beamsplitter}), and ($ii$)
by testing alternatives, such as using grayscale frames reconstructed from the
events (e.g., using state of the art \cite{Rebecq19pami}).  The latter was
tested on DSO and our approach.  The results from DSO are reported in the main
paper, indicated with DSO$^\dagger$ in \cref{tab:compare:frames}.
We observed that it did not produce good results in the bin and
boxes sequences. We observed the same failure effect when using reconstructed
grayscale frames on EDS. Nevertheless, we expect that if using a
higher-end device (e.g., ($i$)) is not an option, better grayscale frames for VO
with our method would be obtained in the near future by means of improvements in
image reconstruction methods and/or event cameras (i.e., lower event noise).
Event cameras are evolving fast, and new prototypes, in combination with
standard sensors may occur in the near future, for example to advance
computational photography (e.g., \cite{Tulyakov21cvpr}).

\subsection{Computational Performance}
The current implementation of our method is un-optimized and it is about
$5\times$ slower than real time.  However, we think that there is large room for
code improvement and engineering to make it real-time capable.  Specifically,
the back-end is un-optimized since automatic differentiation in
Ceres~\cite{ceres-solver} is not real time when having a large number of
parameters (i.e., 4000 points in a 7-keyframe sliding window). 
It could be sped up by feeding fewer points to the back-end and better
selecting a sparse set of points. 
This is the reason why we combined our front-end with DSO's \cite{Engel17pami} back-end,
which is optimized and real-time capable with up to 2000 points in a similar
sliding-window size.
The main purpose of our work is to understand the limitations of previous event-based methods 
(e.g., why do many of the prior works lose track or have large errors?) 
and overcome them with a new design that has not been previously explored. 
Finally, if the method is used offline, e.g., for recovering a scene map 
and/or accurate camera trajectory, runtime is not an issue.

\subsection{Setting an Arbitrary Scale for the World}
Absolute scale is not observable in monocular VO without additional information
or an IMU.  Hence, in our method we need to provide values to set the depth
range of the initial map, which will guide the rest of the visual odometry
process.  We did not consider using an IMU (accelerometer and gyroscope) to
focus on the visual aspects of odometry.  Sensor fusion with an IMU is left as
future work.

\cleardoublepage
{\small
\bibliographystyle{ieeetr_fullname} %
\bibliography{all}
}

\end{document}